\documentclass{svproc}
\usepackage{url}

\usepackage[accsupp]{axessibility}  
\usepackage{algorithm}
\usepackage{algorithmic}
\usepackage{amsfonts}           
\usepackage{amsmath}
\usepackage{amssymb}
\usepackage{array}
\usepackage{booktabs}           
\usepackage[labelfont=bf]{caption}
\usepackage[T1]{fontenc}        
\usepackage{graphicx}
\usepackage{hyperref}           
\usepackage[utf8]{inputenc}     
\usepackage{microtype}          
\usepackage{multirow}
\usepackage{nicefrac}           
\usepackage{pifont}
\usepackage{subcaption}
\usepackage{url}                
\usepackage[framemethod=TikZ]{mdframed}
\usepackage{pgfplots}
\pgfplotsset{compat=1.18}
\usepackage{tabularx}
\usepackage{xcolor}
\usepackage{colortbl}
\usepackage{adjustbox}

\definecolor{darkgreen}{RGB}{0,100,0}

\mdfsetup{
   middlelinecolor=black,
   middlelinewidth=0.5pt,
   backgroundcolor=gray!10,
   innerleftmargin=4pt,
   innerrightmargin=4pt,
   innertopmargin=4pt,
   innerbottommargin=4pt,
   roundcorner=4pt
}

\usepackage{amsmath,amsfonts,bm}
\newcommand{\softmax}{\mathrm{softmax}}
\newcommand{\sC}{\mathbb{C}}
\newcommand{\sM}{\mathbb{M}}
\newcommand{\sP}{\mathbb{P}}
\newcommand{\vx}{\bm{x}}
\newcommand{\vy}{\bm{y}}
\newcommand{\vp}{\bm{p}}
\newcommand{\cmark}{\ding{51}}%
\newcommand{\xmark}{\ding{55}}%
\newcommand{\hb}[1]{\textcolor{blue}{#1}}

\begin{document}
\mainmatter              
\title{Classroom-Inspired Multi-Mentor Distillation with Adaptive Learning Strategies}
\titlerunning{Classroom-Inspired Multi-Mentor Distillation}  
%
\author{Shalini Sarode\inst{1}\inst{2}\thanks{Equal contribution.} \and Muhammad Saif Ullah Khan\inst{1}\inst{2}$^\star$ \and
Tahira Shehzadi \inst{1}\inst{2} \and Didier Stricker \inst{1}\inst{2} \and Muhammad Zeshan Afzal \inst{1}\inst{2}}
\authorrunning{Sarode et al.} 
%
%
\institute{Department of Computer Science, Rhineland-Palatinate Technical University of Kaiserslautern (RPTU), 67663 Kaiserslautern, Germany,\\
\and
Augmented Vision Group, German Research Center for Artificial Intelligence (DFKI), 67663 Kaiserslautern, Germany}

\maketitle              

\begin{abstract}
We propose \textbf{ClassroomKD}, a novel multi-mentor knowledge distillation framework inspired by classroom environments to enhance knowledge transfer between the student and multiple mentors with different knowledge levels. Unlike traditional methods that rely on fixed mentor-student relationships, our framework dynamically selects and adapts the teaching strategies of diverse mentors based on their effectiveness for each data sample. ClassroomKD comprises two main modules: the \textbf{Knowledge Filtering (KF)} module and the \textbf{Mentoring} module. The KF Module dynamically ranks mentors based on their performance for each input, activating only high-quality mentors to minimize error accumulation and prevent information loss. The Mentoring Module adjusts the distillation strategy by tuning each mentor's influence according to the dynamic performance gap between the student and mentors, effectively modulating the learning pace. Extensive experiments on image classification (CIFAR-100 and ImageNet) and 2D human pose estimation (COCO Keypoints and MPII Human Pose) demonstrate that ClassroomKD outperforms existing knowledge distillation methods for different network architectures. Our results highlight that a dynamic and adaptive approach to mentor selection and guidance leads to more effective knowledge transfer, paving the way for enhanced model performance through distillation.
\keywords{knowledge distillation, multi-mentors, lifelong learning, image classification, pose estimation, classroom learning}
\end{abstract}
\section{Introduction}
\label{sec:intro}

Knowledge distillation (KD)~\cite{hinton2015distilling} is a widely adopted model compression technique in deep learning, where a smaller, more efficient student model learns to replicate the behavior of a larger, more complex teacher model. While traditional KD methods~\cite{hinton2015distilling}\cite{zagoruyko2016paying}\cite{adriana2015fitnets} typically employ a single teacher, multi-teacher (or multi-mentor) distillation has been proposed to further enhance performance by leveraging an ensemble of teachers~\cite{you2017learning}. This setup is expected to provide richer and more diverse knowledge, improving the student’s generalization and robustness. We use the term \textbf{mentor} to describe all networks involved in teaching the student, regardless of their size or role.

\begin{figure}[ht] 
    \centering
    \includegraphics[width=0.9\linewidth]{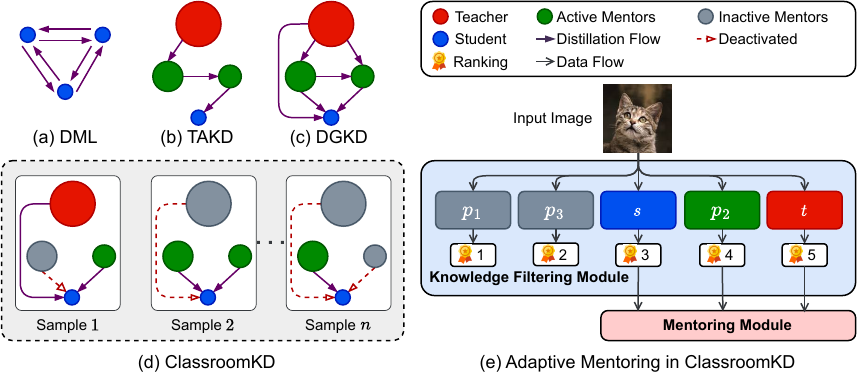}
    \caption{(a) DML: Peer models learn from each other without a hierarchical teacher structure. (b) TAKD: A sequential mentor-student hierarchy with large-to-small knowledge transfer. (c) DGKD: Each mentor teaches all smaller models. (d) ClassroomKD: Our proposed method dynamically selects mentors for each data sample based on the current input and ranks them using the Knowledge Filtering Module. (e) Adaptive Mentoring: The Mentoring Module adjusts teaching strategies of each active mentor according to dynamic rankings, ensuring optimal knowledge transfer.}
    \label{fig:kd_intro}
\end{figure}

Despite its potential benefits, multi-mentor distillation faces several significant challenges:

\textbf{Large Capacity Gap}: Employing multiple large mentors can create a substantial capacity gap between the collective representation power of the mentors and that of the student. This gap can hinder the student’s ability to effectively mimic the combined knowledge of the mentors, leading to suboptimal learning outcomes. To bridge this gap, some works~\cite{mirzadeh2019improved,son2021densely} have introduced intermediate-sized mentors alongside a large teacher. However, smaller mentors may be less effective, potentially introducing additional errors into the student's knowledge. 

\textbf{Error Accumulation}: The lower performance of smaller mentors can contribute to cumulative errors in the distillation process. This is particularly problematic in sequential distillation frameworks like TAKD (Figure~\ref{fig:kd_intro}(b)), where each mentor teaches only the subsequent smaller model. Such setups can lead to an "error avalanche," where inaccuracies from lower-performing mentors degrade the student’s performance~\cite{son2021densely}. Although DGKD (Figure~\ref{fig:kd_intro}(c)) attempts to mitigate this by allowing each mentor to teach all smaller models and randomly dropping some mentors during training, these strategies can result in valuable information loss and reduced learning efficiency.

\textbf{Lack of Dynamic Adaptation}: The performance gap between the student and its mentors is not static; it evolves throughout training (as visualized in Appendix~\ref{sec:dynamic-capacity-gap} Figures~\ref{fig:softmax_ep1}-\ref{fig:softmax_ep228}). Current methods do not adequately address these dynamic scenarios, limiting the effectiveness of multi-mentor distillation~\cite{hao2024one}. Without an adaptive strategy, the potential benefits of multi-mentor distillation are not fully realized.

Observing that \textbf{(1)} a mentor's performance varies across different data samples, \textbf{(2)} each mentor possesses distinct teaching capabilities due to varying capacity gaps, and \textbf{(3)} the performance gap evolves during training, we draw parallels between knowledge distillation and Vygotsky’s Zone of Proximal Development (ZPD) (1978). His theory emphasizes learning with a More Knowledgeable Other (MKO) and the need for scaffolded support. 

We propose \textbf{ClassroomKD} (Figure~\ref{fig:kd_intro}(d)), a novel multi-mentor distillation framework inspired by classroom dynamics (see Appendix~\ref{sec:survey}). Our method introduces two key modules (Figure~\ref{fig:kd_intro}(e)) designed to address the following questions:

\begin{mdframed}
\textbf{Q1: Which mentors are effective teachers for a given data sample?}

We introduce the \textbf{Knowledge Filtering Module} to intelligently select mentors(or the \textit{MKOs}). This module dynamically ranks all mentors based on their performance for each input, activating only those with sufficient performance. A mentor is deemed effective and activated if its predictions are accurate and more confident than the student’s. This minimizes error accumulation and information loss.
\end{mdframed}

\begin{mdframed}
\textbf{Q2: How much information should the student learn from each active mentor?}

Our \textbf{Mentoring Module} addresses this by tuning the teaching strategy based on the performance gap between the student and each active mentor. Specifically, we adjust each mentor's distillation temperature to control the teaching pace(\textit{scaffolding}), allowing the student to appropriately weigh information received from each mentor before integrating it into its own knowledge.
\end{mdframed}

By addressing these questions iteratively, ClassroomKD ensures a continuously optimized learning process that adapts to the student’s evolving capabilities. Our \textbf{contributions} are as follows:

\begin{enumerate}
    \item \textbf{ClassroomKD Framework}: We introduce ClassroomKD, a novel multi-mentor distillation framework to dynamically select effective mentors and adapt teaching strategies.
    \item \textbf{Knowledge Filtering Module}: We develop a Knowledge Filtering Module to enhance distillation quality by selectively activating high-performance mentors, thereby reducing error accumulation and preventing information loss.
    \item \textbf{Mentoring Module}: We create a Mentoring Module that dynamically adjusts teaching strategies based on the performance gap between the student and each active mentor, optimizing the knowledge transfer process.
    \item \textbf{Empirical Validation}: Through extensive experiments on image classification (CIFAR-100 and ImageNet) and 2D human pose estimation (COCO Keypoints and MPII Human Pose), we demonstrate that ClassroomKD significantly outperforms state-of-the-art KD methods.
\end{enumerate}

\section{Related Work}
\label{sec:related work}

\subsection{Knowledge Distillation Approaches}
Knowledge distillation (KD)~\cite{hinton2015distilling} is a widely adopted technique for compressing deep neural networks, where a smaller student model learns from a larger teacher model by minimizing the distance between their output probability distributions, or soft labels. Traditional KD methods primarily focus on \textbf{logit-based distillation}, where the student learns directly from the teacher's output logits. Notable methods include PKT~\cite{passalis2018learning}, which employs probabilistic knowledge transfer, FT~\cite{kim2018paraphrasing}, which transfers factorized feature representations, and AB~\cite{heo2019knowledge}, which leverages activation boundaries formed by hidden neurons.

\textbf{Feature-based distillation} methods transfer knowledge by aligning intermediate representations between the teacher and student. FitNets~\cite{adriana2015fitnets} introduced this approach using intermediate feature maps for training. Later methods like AT~\cite{zagoruyko2016paying}, VID~\cite{ahn2019variational}, and CRD~\cite{tian2020contrastive} enhance knowledge transfer by matching attention maps, utilizing variational information distillation, and employing contrastive learning, respectively.

\textbf{Relation-based methods} focus on preserving the structural relationships within the teacher's feature maps. RKD~\cite{park2019relational} maintains data point structures through relational knowledge distillation, while SP~\cite{tung2019similarity} and SRRL~\cite{yang2021knowledge} optimize for similarity-preserving objectives. DIST~\cite{huang2022knowledge} addresses large capacity gaps by applying a correlation-based loss to maintain both inter-class and intra-class relationships, enhancing distillation efficiency.

Recent approaches have explored more specialized distillation techniques. WSLD~\cite{zhou2021rethinking} introduces weighted soft labels to balance bias-variance trade-offs, while One-to-All Spatial Matching KD~\cite{lin2022knowledge} focuses on spatial matching techniques. OFA~\cite{hao2024one} optimizes feature-based KD by projecting features onto the logit space, significantly improving performance for heterogeneous models. To enhance distillation effectiveness, several methods have incorporated adaptive strategies. CTKD~\cite{li2023curriculum} dynamically learns the temperature during training to gradually increase learning difficulty, and DTKD~\cite{wei2024dynamic} employs real-time temperature scaling to improve knowledge transfer efficiency.

\subsection{Multi-Teacher Knowledge Distillation}
Multi-teacher distillation methods aim to further enhance student performance by leveraging an ensemble of mentors~\cite{you2017learning}.

\textbf{Online knowledge distillation} has been particularly successful in this context. Deep Mutual Learning (DML)~\cite{zhang2018mutual} introduces a framework where multiple peer models learn from each other simultaneously during training, fostering collaborative learning among smaller networks and outperforming traditional one-way (offline) distillation. Other online methods include ONE~\cite{zhu2018knowledge}, OKDDip~\cite{chen2020online}, and FFM~\cite{li2020online}, which often outperform offline methods. Online distillation has also been extended to pose estimation tasks~\cite{kd_pose7}. SHAKE~\cite{li2022shadow} proposed using proxy teachers with shadow heads to use the benefits of online distillation in offline settings.

To address the \textbf{capacity gap} in multi-teacher setups, Teacher-Assistant KD (TAKD)~\cite{mirzadeh2019improved} employs intermediate-sized teacher assistants (TAs) to bridge the gap between the largest teacher and the student. However, sequential distillation through TAs can result in an "error avalanche", where errors propagate at each step, reducing final performance. Adaptive Ensemble Knowledge Distillation (AEKD)~\cite{du2020agree} mitigates this issue by using an adaptive dynamic weighting strategy to reduce error propagation in the gradient space. Densely Guided KD (DGKD)~\cite{son2021densely} further improves upon these methods by guiding each TA with both larger TAs and the main teacher, enabling a more gradual and effective transfer of knowledge. Additionally, DGKD introduces a strategy of randomly dropping mentors during training to expose the student to diverse learning sources, enhancing overall learning robustness.

While existing multi-teacher methods offer various mechanisms for knowledge distillation, they still grapple with challenges such as managing the capacity gap, mitigating error accumulation, and adapting to dynamic mentor-student relationships. 

\begin{figure}[ht]
    \centering
    \includegraphics[width=\linewidth]{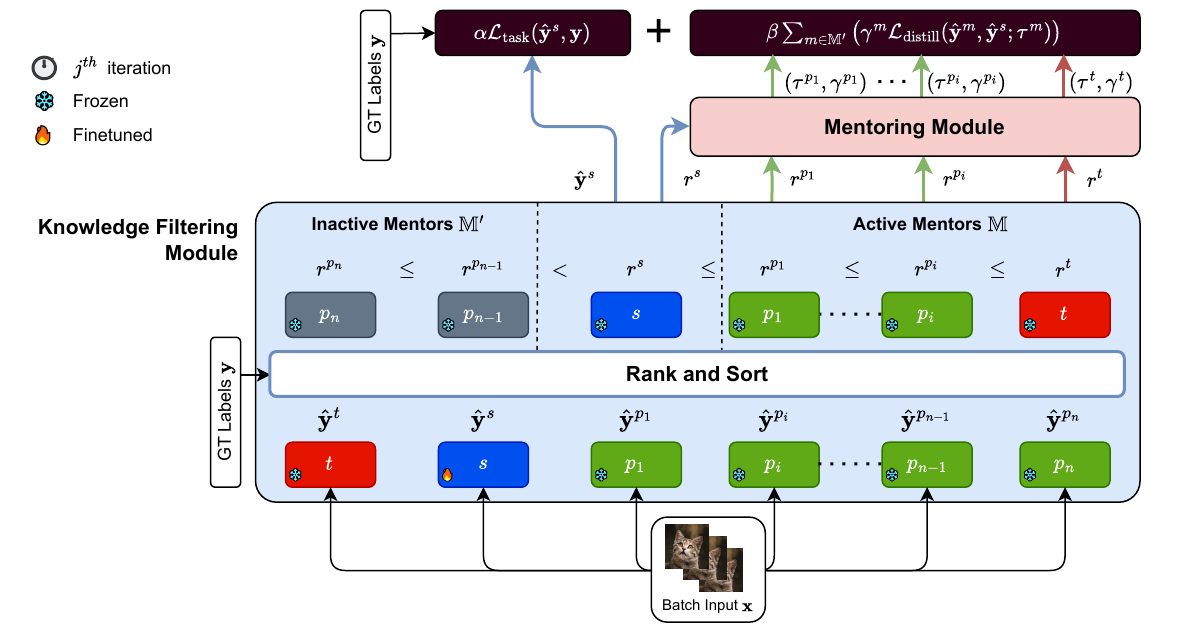}
    \caption{\textbf{The ClassroomKD framework.} comprises a \textbf{Knowledge Filtering (KF) Module} and a \textbf{Mentoring Module}. The KF Module optimizes learning by selectively incorporating feedback from higher-ranked mentors, reducing noise transfer and preventing error accumulation. The Mentoring Module adjusts mentor influence based on their performance relative to the student.}
    \label{fig:kd_main}
\end{figure}

\section{Methodology}
\label{sec:methodology}

ClassroomKD is a novel multi-mentor distillation framework inspired by real-world classroom environments. It is designed to address the challenges of large capacity gaps, error accumulation, and lack of dynamic adaptation. Our framework is illustrated in Figure~\ref{fig:kd_main}.

\textbf{Classroom Definition.} A classroom comprises \textbf{(1)} a high-capacity \textit{teacher} model, \(t\), \textbf{(2)} a small \textit{student} model, \(s\), and \textbf{(3)} \(n\) \textit{peer} models of intermediate capacities, \(\sP = \{p_i\}_{i=1}^n\). We define \(\sM = \{t\} \cup \sP\) as the set of pre-trained mentors that remain frozen during the student’s training process. At each training step, the student distills knowledge from a dynamically selected subset of mentors, called the \textit{active mentors} (\(\sM' \subseteq \sM\)). The set of all classroom models is denoted \(\sC = \{s\} \cup \sM\). We use the Knowledge Filtering (KF) Module for intelligent mentor selection and the Mentoring Module to adjust the teaching pace based on the capacity gap of each mentor-student pair.

\subsection{Knowledge Filtering Module}

The KF Module is designed to intelligently select which mentors should contribute to the student’s learning process for each data sample. This selective approach mitigates error accumulation and prevents the student from learning from less effective mentors.

Let \(\vx = \{x_k\}_{k=1}^N\) be a batch of training data with size \(N\), and \(\vy = \{y_k\}_{k=1}^N\) be the ground-truth labels. The batch inputs \(\vx\) are forwarded through all classroom models to obtain the predicted logits \(\hat{\vy}^m\), which are then converted to probabilities with a softmax operation. We isolate the probability assigned to the true class \(\vy\) and compute a weighted average of the correct prediction probability across the batch for each model. For all \(m \in \sC\), this is defined as:
\begin{align}
    \hat{\vy}^m &= m(\vx) \\
    \vp^m &= \softmax(\hat{\vy}^m)  \\
    \vp^m_{\text{gt}} &= \vp^m[\vy] = 1/(\text{exp}(\mathrm{CELoss}(\hat{\vy}^m, \text{targets})) \label{eq:rank_gt}\\
    w^m &= \frac{1}{N} \sum_{k=1}^N \vp^m_{\text{gt}}(x_k)
\end{align}
The weights \(w^m\) reflect the performance of model \(m\) on the current training batch. We use the computed weights as a proxy for mentor suitability in the distillation process and rank mentors based on their relative performance to all classroom models:
\begin{align} 
    r^m &= \lambda \left(\frac{w^m}{\sum_{m\in\sC} {w^m}}\right)
    \label{eq:rank_score}
\end{align} 
where \(r^m\) is a normalized ranking score of model \(m\), and \(\lambda\) is a scaling parameter set to the number of mentors in the classroom. Active mentors \(\sM'\) are defined as those with higher ranks than the student:
\begin{align}
    \sM' &= \{ m \mid m \in \sM \text{ and } r^m > r^s \}
\end{align}
This ensures the student learns from high-quality sources by selecting mentors based on their performance ranks. This selective approach \textbf{prevents error accumulation} as only mentors outperforming the student can teach it, avoiding the propagation of errors from less effective mentors. Additionally, it \textbf{avoids information loss} by consistently selecting the best-performing mentors, unlike random mentor-dropping strategies~\cite{son2021densely}.

\subsection{Mentoring Module}

The Mentoring Module dynamically adjusts the influence of each active mentor based on the mentor-student performance gap. This \textbf{adaptive teaching strategy} facilitates effective knowledge transfer tailored to the student's evolving ability to absorb information from each mentor.

The distillation loss minimizes KL divergence between the student and mentor's output distributions:
\begin{equation}
    \mathcal{L}_{distill}(P, Q; \tau) = \tau^2 \cdot \text{KL}\left(\softmax\left({P}/{\tau}\right) \,\|\, \softmax\left({Q}/{\tau}\right)\right)
\end{equation} where \(P\) and \(Q\) represent the logits from the mentor and student networks, respectively, and \(\tau\) is a temperature hyperparameter that smooths the probability distributions during the distillation process.

The temperature \(\tau\) controls the sharpness of the probability distributions, affecting the knowledge transfer from a mentor to the student. For each active mentor \(m \in \sM'\), we adjust the distillation temperature \(\tau^m\) based on the performance gap between the student and the mentor. The performance gap is measured as the difference in their ranking scores:
\begin{align}
    \label{eq:tau_t}
    \Delta r^m &= |r^m - r^s| / r^m \\
    \tau^m     &= 1 + \Delta r^m \cdot \tau
    \label{eq:tau_adjustment}
\end{align}
Here, \(\tau\) is the base temperature, and \(\tau^m\) increases with \(\Delta r^m\), which represents the mentor-student performance gap. A larger \(\Delta r^m\) results in a higher \(\tau^m\), smoothing the mentor's output distribution. This adjustment theoretically slows down the distillation process by softening the mentor's predictions, allowing the student to assimilate knowledge more gradually when the performance gap is large. Conversely, the student receives sharper, more direct guidance when the gap is small.

The total loss \(\mathcal{L}\) is computed by combining a task-specific loss $\mathcal{L}_{\mathrm{task}}$ with the weighted distillation losses from all active mentors:
\begin{align} 
\mathcal{L}_{\mathrm{classroom}} &= \alpha \mathcal{L}_{\mathrm{task}}(\hat{\vy}^s, \vy) + \beta \sum_{m \in \sM'} \gamma^m \mathcal{L}_{\mathrm{distill}}(\hat{\vy}^m,\hat{\vy}^s;\tau^m) \\
\mathcal{L} &= \delta(\mathcal{L}_{\mathrm{task}}(\hat{\vy}^s, \vy) + \mathcal{L}_{\mathrm{distill}}(\hat{\vy}^t,\hat{\vy}^s;\tau^t=1)) + \mathcal{L}_{\mathrm{classroom}}
\label{eq:total_loss} 
\end{align}
Here, \(\alpha = r^s\) represents the student's self-confidence, which scales the task-specific loss. As the student’s rank \(r^s\) improves, \(\alpha\) increases, encouraging the student to rely more on its own predictions. For each mentor \(m\), \(\gamma^m = r^m\) scales the corresponding distillation loss, where \(r^m\) is the mentor's rank relative to the student. \(\beta\) is a hyperparameter to control the influence of distillation loss relative to the task loss. This weighing, along with the mentor-specific temperature \(\tau^m\), ensures that higher-performing mentors have a greater influence on the student's learning, with each mentor distilling knowledge at an appropriate rate based on the performance gap. We use Cross-Entropy Loss for classification and MSE Loss for pose estimation tasks.

This promotes independent learning by increasing the student's reliance on its own task performance as its confidence grows. It also ensures that the student benefits from guidance based on the relative performance of the active mentors, effectively balancing task-specific training with distillation from the most suitable mentors. This dynamic and adaptive approach ensures \textbf{optimized knowledge transfer}, minimizes error accumulation, and enhances the overall performance of the student model.

\section{Experiments}
\label{sec:experiments}

This section presents our experiments to evaluate the effectiveness of ClassroomKD using different datasets. We primarily use CIFAR-100~\cite{krizhevsky2009learning} classification for detailed comparisons with state-of-the-art single and multiple-teacher distillation methods. This also includes online approaches using multiple mentors. In addition, we also report results on ImageNet~\cite{deng2009imagenet} classification and human pose estimation using the COCO Keypoints~\cite{lin2014microsoft} and MPII Human Pose~\cite{andriluka20142d} datasets. Our results show that ClassroomKD outperforms existing methods under various settings, highlighting the robustness and adaptability of our method.

\noindent
\begin{minipage}{0.6\linewidth}
\textbf{Implementation Details.} For CIFAR-100, we train for 240 epochs with a batch size of 64, a learning rate of 0.05 decayed by 10\% every 30 epochs, and a 120-epoch warm-up phase. We use SGD with 0.9 momentum and \(5 \times 10^{-4}\) weight decay. The temperature \(\tau\) is set to 12 via grid search (Figure~\ref{fig:temperature-search}). For ImageNet, models are trained for 100 epochs with \(\tau = 4\). For pose datasets, models are trained for 210 epochs with \(\tau = 4\). The scaling factor \(\lambda\) is \(n+1\) for all experiments, where \(n\) is the number of peers. We used \(\beta=1.0\) for classification and \(\beta=2.5\) for pose estimation. Furthermore, we set \(\delta=0\) for CIFAR100 dataset and \(\delta=1\) for the large-scale Imagenet dataset.
\end{minipage}
\hspace{0.03\linewidth}
\begin{minipage}{0.36\linewidth}
    \begin{tikzpicture}
        \begin{axis}[
            scale only  axis,
            width=0.7\linewidth,
            height=2cm,
            xlabel={Temperature},
            ylabel={Student Top-1},
            xlabel near ticks,
            ylabel near ticks,
            xtick={2, 4, 6, 8, 10, 12},
            grid=major,
            grid style={dotted,color=black},
            legend pos=north west,
            label style={font=\scriptsize},
            legend style={font=\scriptsize},
            tick label style={font=\scriptsize},
        ]
    
        \addplot[
            color=blue, mark=square*, mark options={scale=1.2}, thick
        ] coordinates {
            (2, 65.87)
            (4, 65.55)
            (6, 65.58)
            (8, 65.72)
            (10, 65.43)
            (12, 65.96)
        };
        
    
        \end{axis}
    \end{tikzpicture}
    \captionof{figure}{\textbf{Temperature selection.} Grid search using fixed-temperature KD, with the best student performance at $\tau=12$, used as the base temperature in Eq.~\ref{eq:tau_adjustment}.}
    \label{fig:temperature-search}
\end{minipage}

We follow standard training protocols, with mentors pre-trained and kept frozen.

\subsection{CIFAR-100 Classification}

We begin by comparing ClassroomKD against single-teacher distillation methods in Table~\ref{tab:classification-cifar100}. The table includes a variety of teacher–student pairs, covering both homogeneous (e.g., ResNet 110 → ResNet 20) and heterogeneous (e.g., VGG 13 → MobileNetV2) architectures.

\begin{table}[ht]
    \centering
    \scriptsize
    \caption{\textbf{Comparison with single-teacher distillation methods on CIFAR-100 classification.} We report top-1 accuracy (\%). KD methods are grouped by feature, relation, and logit-based. Best values in logit-based methods are \textbf{bold}, second-best \underline{underlined}, and overall best \hb{blue}}
    \label{tab:classification-cifar100}
    \begin{tabularx}{\linewidth}{Xlcccccccc}
        \toprule
        Method &
        \multicolumn{4}{c}{Homogeneous architectures} &
        \multicolumn{5}{c}{Heterogeneous architectures} \\
        \cmidrule(lr){1-1}                   \cmidrule(lr){2-5}              \cmidrule(lr){6-10}
        Teacher                          & R110  & R110  & R56   & VGG13 & VGG13 & R32×4 & W-40x2& R50   & Swin-T \\
        Student                          & R20   & R32   & R20   & VGG8  & MBV2 & SN-V2 & SN-V1 & MBV2 & R18 \\

        \cmidrule(lr){1-1}                   \cmidrule(lr){2-5}              \cmidrule(lr){6-10}
        NOKD                              & 69.06 & 71.14 & 69.06 & 70.68 & 64.60 & 71.82 & 70.50 & 64.60 & 74.01 \\

        \cmidrule(lr){1-1}                   \cmidrule(lr){2-5}              \cmidrule(lr){6-10}
        FitNets~\cite{adriana2015fitnets}& 68.99 & 71.06 & 69.21 & 73.54 & 64.14 & 73.54 & 73.73 & 63.16 & 78.87 \\
        AT~\cite{zagoruyko2016paying}    & 70.22 & 72.31 & 70.55 & 73.62 & 59.40 & 72.73 & 73.32 & -     & -     \\
        VID~\cite{ahn2019variational}    & 70.16 & 72.61 & 70.38 & 73.96 & -     & 73.40 & 73.61 & 67.57 & -     \\  
        CRD~\cite{tian2020contrastive}   & 71.46 & 73.48 & 71.16 & 73.94 & 69.73 & 75.65 & 76.05 & 69.11 & 77.63 \\
        SimKD~\cite{chen2022knowledge}   & -     & -     & -     & 74.93 & -     & \hb{77.49} & -     & -     & -     \\  
        SMKD~\cite{lin2022knowledge}     & 71.70 & 74.05 & 71.59 & 74.39 & -     & -     & -     & -     & -     \\

        \cmidrule(lr){1-1}                   \cmidrule(lr){2-5}              \cmidrule(lr){6-10}
        RKD~\cite{park2019relational}    & 69.25 & 71.82 & 69.61 & 73.72 & 64.52 & 73.21 & 72.21 & 64.43 & 74.11 \\
        SP~\cite{tung2019similarity}     & 70.04 & 72.69 & 69.67 & 73.44 & 66.30 & 74.56 & 74.52 & -     & -     \\
        SRRL~\cite{yang2021knowledge}    & 71.51 & 73.80 & -     & 73.23 & 69.34 & 75.66 & \hb{76.61} & -     & -     \\
        DIST~\cite{huang2022knowledge}   & -     & -     & 71.75 & -     & -     & 77.35 & -     & 68.66 & 77.75 \\

        \cmidrule(lr){1-1}                   \cmidrule(lr){2-5}              \cmidrule(lr){6-10}
        KD~\cite{hinton2015distilling}   & 70.67 & 73.08 & 70.66 & 72.98 & 67.37 & 74.45 & 74.83 & 67.35 & 78.74 \\
        PKT~\cite{passalis2018learning}  & 70.25 & 72.61 & 70.34 & 73.37 & -     & 74.69 & 73.89 & 66.52 & -     \\
        FT~\cite{kim2018paraphrasing}    & 70.22 & 72.37 & 69.84 & 73.42 & -     & 72.50 & 72.03 & -     & -     \\
        AB~\cite{heo2019knowledge}       & 69.53 & 70.98 & 69.47 & \underline{74.27} & -     & 74.31 & 73.34 & -     & -     \\
        WSLD~\cite{zhou2021rethinking}   & \textbf{\hb{72.19}} & \underline{74.12} & \textbf{\hb{72.15}} & -     & -     & 75.93 & \underline{76.21} & -     & -     \\
        CTKD~\cite{li2023curriculum}     & 70.99 & 73.52 & 71.19 & 73.52 & 68.46 & 75.31 & 75.78 & 68.47 & -     \\
        DTKD~\cite{wei2024dynamic}       & -     & 74.07 & 72.05 & 74.12 & \underline{69.01} & \underline{76.19} & \textbf{76.29} & \underline{69.10} & -     \\
        OFA~\cite{hao2024one}            & -     & -     & -     & -     & -     & -     & -     & -     & \textbf{\hb{80.54}} \\  

        \cmidrule(lr){1-1}                   \cmidrule(lr){2-5}              \cmidrule(lr){6-10}
        \rowcolor{gray!25}
        Ours                     & \underline{72.06} & \textbf{\hb{74.71}} & \underline{72.13} & \textbf{\hb{75.29}} & \textbf{\hb{70.26}} & \textbf{76.74} & 75.81 & \textbf{\hb{70.23}} &  \underline{80.32}\\
        
        \bottomrule
    \end{tabularx}
\end{table}

\paragraph{Overall Improvements.} ClassroomKD consistently outperforms baseline logit-based methods, as well as many feature-based and relation-based methods. Notably, our method competes favorably with recent approaches that employ adaptive temperature scaling, such as CTKD~\cite{li2023curriculum} and DTKD~\cite{wei2024dynamic}. These gains suggest that our combination of selective mentor activation and dynamically adjusted temperatures more robustly handles the evolving capacity gap than strategies that only tune temperature globally.

\paragraph{Capacity Gap Mitigation.} In large-teacher/small-student pairings (e.g., ResNet-110 → ResNet-20, VGG-13 → MobileNetV2), the capacity gap is significant. ClassroomKD explicitly tackles this by filtering out under-performing mentors in a data-dependent way, reducing “noisy guidance.” This proves especially helpful in preventing a performance plateau observed in many other KD methods when the teacher is much larger.

\subsection{CIFAR-100 Classification with Multiple Mentors}

We next evaluate the multi-mentor scenario in Table~[\ref{tab:classification-cifar100-multiple}], where each classroom includes a single large teacher and several intermediate-capacity peers (details in Appendix~\ref{sec:mentor-configuration}). We compare both online frameworks (e.g., DML~\cite{zhang2018mutual} and SHAKE~\cite{li2022shadow} and offline ones (e.g., AEKD~\cite{du2020agree}, DGKD~\cite{son2021densely}).

\paragraph{Defining a Simple Baseline.} Following SOTA methods~\cite{li2022shadow,zhang2022confidence}, we use \textbf{AVER} as the simplest baseline in our multi-mentor comparisons. This is a direct counterpart of KD in single-teacher experiments and is defined as:
\begin{equation} \mathcal{L}_{\mathrm{AVER}} = \mathcal{L}_{\mathrm{task}}(\hat{\vy}^s, \vy) + \sum_{m \in \sM} \mathcal{L}_{\mathrm{distill}}(\hat{\vy}^m,\hat{\vy}^s;\tau)
\end{equation}
Each teacher is weighted equally without any ranking or temperature adaption; the student naively attempts to learn the aggregate of all teachers' knowledge.

\begin{table}[ht]
\caption{\textbf{Comparison with multi-teacher distillation methods.}}
\label{tab:multi-mentor-results}
\begin{minipage}{0.6\linewidth}
\begin{subtable}[t]{\linewidth}
    \centering
    \scriptsize
    \caption{\textbf{Results on CIFAR-100 classification.} We report top-1 accuracy (\%). KD methods are grouped by online and offline. ClassroomKD is offline. Best and second-best values in offline methods are \textbf{bold} and underlined, respectively, and overall best in \hb{blue}. Complete classroom configurations with details about the peers are provided in the appendix.}
    \label{tab:classification-cifar100-multiple}
    \begin{tabularx}{\linewidth}{Xcccccccccc}
        \toprule
        Method &
        \multicolumn{4}{c}{Same Archs} &
        \multicolumn{2}{c}{Mixed Archs} \\
        \cmidrule(lr){1-1} \cmidrule(lr){2-5} \cmidrule(lr){6-7}
        Teacher                           & WR40x2 & R110  & R56   & VGG13 & VGG13 & W-40x2\\
        Student                           & WR16x2 & R20   & R20   & VGG8  & MBV2 & SN-V1 \\

        \cmidrule(lr){1-1} \cmidrule(lr){2-5} \cmidrule(lr){6-7}
        NOKD                              & 73.64 & 69.06 & 69.06 & 70.68 & 64.60 & 70.50 \\

        \cmidrule(lr){1-1} \cmidrule(lr){2-5} \cmidrule(lr){6-7}
        DML        & 74.83 & 70.55 & 70.24 & 72.86 & 66.30 & 74.52 \\
        ONE      & 74.68 & 70.77 & 70.43 & 72.01 & 66.26 & -     \\
        SHAKE        & 75.78 & -     & 71.62 & 73.85 & 68.81 & 76.42 \\

        \cmidrule(lr){1-1} \cmidrule(lr){2-5} \cmidrule(lr){6-7}
        TAKD & 75.04 & -     & 70.77 & 73.67 & -     & -     \\
        AEKD          & 75.68 & \underline{71.36} & 71.25 & \underline{74.75} & 68.39 & 76.34 \\
        EBKD & & & & 74.10 & 68.24 & \underline{76.61} \\
        DGKD       & \underline{76.24} & -     & \underline{71.92} & 74.40 & -     & -     \\
        CA-MKD & - & - & - & 74.30 & \underline{69.41} & \textbf{\hb{77.94}} \\
        \cmidrule(lr){1-1} \cmidrule(lr){2-5} \cmidrule(lr){6-7}
        AVER                              & 74.98 & 71.20 & 71.08 & 73.18 & 62.94 & 73.00 \\
        \rowcolor{gray!25}
        Ours                  & \textbf{\hb{76.74}}  & \textbf{\hb{72.06}} & \textbf{\hb{72.13}} & \textbf{\hb{75.29}} & \textbf{\hb{70.26}} & 75.81 \\
        \bottomrule
    \end{tabularx}
\end{subtable}
\end{minipage}
\hfill
\begin{minipage}{0.38\linewidth}
\begin{subtable}[t]{\linewidth}
    \centering
    \scriptsize

    \caption{\textbf{Results on ImageNet.}}
    \label{tab:result_imagenet}
     \begin{tabularx}{\linewidth}{Xc}
        \toprule
        \multicolumn{2}{c}{T: R34, S: R18, 4 P} \\
        \midrule
        NOKD    & 69.75 \\
        DML     & 71.03 \\
        ONE     & 70.55 \\
        SHAKE   & \textbf{72.07} \\
        KD      & 70.66 \\
        CRD     & 71.17 \\
        AVER    & 70.63 \\
        \cellcolor{gray!25} Ours &  \cellcolor{gray!25} \underline{71.49}\\
        \bottomrule
    \end{tabularx}
\end{subtable}
\begin{subtable}[t]{\linewidth}
    \centering
    \scriptsize
    \caption{\textbf{Pose Estimation Results} with 4 mentors. We report PCKh for MPII and AP for COCO.}
    \label{tab:pose_results}
    \begin{tabularx}{\linewidth}{Xccc}
        \toprule
        Dataset & \multicolumn{2}{c}{MPII} & COCO \\
        \cmidrule(lr){1-1}
        \cmidrule(lr){2-3}
        \cmidrule(lr){4-4}
        Teacher & \multicolumn{2}{c}{HRNet-W32-D} & RTMP-L \\
        Student & \multicolumn{2}{c}{LiteHRNet-18} & RTMPose-t \\
        Peers & Same & Mixed & Same \\
        \cmidrule(lr){1-1}
        \cmidrule(lr){2-2}
        \cmidrule(lr){3-3}
        \cmidrule(lr){4-4}
        NOKD  & 85.91 & 85.91 & 68.20 \\
        AVER  & 86.64 & 86.07 & 69.26 \\
        \rowcolor{gray!25}
        Ours  & \textbf{86.72} & \textbf{86.37} & \textbf{69.73} \\
        \bottomrule
    \end{tabularx}
\end{subtable}
\end{minipage}
\end{table}

Directly aggregating multiple mentors (AVER) often yields modest improvements. ClassroomKD takes this further by ranking and selectively activating mentors, reducing the risk of error accumulation from weaker ones.

\paragraph{Comparison with Specialized Methods.} Techniques like TAKD~\cite{mirzadeh2019improved} or DGKD~\cite{son2021densely} were devised primarily to address large capacity gaps and error propagation, yet ClassroomKD still shows consistently higher accuracy. This underscores the advantage of our fine-grained, per-sample mentor filtering over either purely sequential or random-drop strategies.

\paragraph{Fewer Mentors, Stronger Gains.} Interestingly, we outperform approaches like AEKD~\cite{du2020agree}, which rely on more mentors than we do. This highlights that mentor \emph{quality} and selective usage are more crucial to final student performance than simply increasing the number of possible teachers.

\subsection{ImageNet Classification with Multiple Mentors}

To assess scalability, we evaluate ClassroomKD on ImageNet in Table~\ref{tab:result_imagenet}. ClassroomKD maintains its improvements even when dealing with large-scale data, demonstrating the generality of the rank-based mentor selection. While online SHAKE remains slightly higher, our method still surpasses classic offline KD methods and the naive multi-mentor baseline (AVER).

\paragraph{Partial Online vs. Offline Gap.} Our offline approach does not achieve the same final result as online SHAKE, which benefits from constant inter-model updating. Still, we remain close, suggesting that an adaptive offline framework can approximate or rival online methods without overheads such as co-training multiple large models simultaneously.

\subsection{Pose Estimation with Multiple Mentors}

Finally, we test ClassroomKD on 2D human pose estimation tasks—both on COCO Keypoints and MPII Human Pose—presented in Table~\ref{tab:pose_results}. Each classroom includes a large high-accuracy teacher (e.g., HRNet-W32-D) plus up to four additional peers. Additional details on our adaptation for pose estimation (e.g., how we compute ranks for heatmap vs. SimCC heads) appear in Appendix~\ref{sec:pose_estimation}.

\paragraph{Performance Gains.} Unlike classification, pose estimation requires learning structured output (e.g., heatmaps, SimCC x/y logits). Our experiments show that the same rank-based selection and adaptive temperature scaling indeed transfer effectively to these more complex output heads. ClassroomKD consistently achieves better PCKh (MPII) and AP (COCO) than the simplest multi-mentor baseline (AVER). This is mainly attributed to removing guidance from less reliable peers—particularly at the early epochs when the capacity gap is large.

\section{Ablation Studies}

We conduct a series of ablation studies to understand the individual contributions of different components of our ClassroomKD framework, providing insights into our design choices.

\begin{table*}[ht]
    \centering
    \scriptsize
    \caption{\textbf{Ablation study} to assess the impact of ClassroomKD components.}
    \begin{subtable}[t]{0.5\linewidth}
        \centering
        \scriptsize
        \caption{\textbf{Role of Multiple Mentors.} Single-teacher distillation slightly improves student performance compared to vanilla training. Intermediate mentors (peers) and adaptive distillation further enhances learning.}
        \label{tab:mentor-role}
        \begin{tabularx}{\linewidth}{ccccc}
            \toprule
            Student                                    & Teacher                 & Peers                   & Adaptive & Top-1 Accuracy \\
            \midrule
            \color{black}\cmark                    & \color{black}\xmark     & \color{black}\xmark     & \color{black}\xmark     & 63.31 \\
            \color{black}\cmark                    & \color{black}\cmark & \color{black}\xmark     & \color{black}\xmark     & 63.35 \\
            \color{black}\cmark & \color{black}\cmark & \color{black}\cmark & \color{black}\xmark     & 65.96 \\
            \rowcolor{gray!25} \color{black}\cmark & \color{black}\cmark & \color{black}\cmark & \color{black}\cmark & \textbf{68.52} \\
            \bottomrule
        \end{tabularx}
    \end{subtable}
    \hfill
    \begin{subtable}[t]{0.48\linewidth}
        \centering
        \scriptsize
        \caption{\textbf{Adaptive Distillation in ClassroomKD.} We analyze the role of the KF Module and Mentoring Module in our adaptive method. Both components contribute to overall performance.}
        \label{tab:classroom-modules}
        \begin{tabularx}{\linewidth}{ccc}
            \toprule
            KF Module & Mentoring Module & Top-1 Accuracy \\
            \midrule
            \color{black}\xmark                        & \color{black}\xmark     & 65.96 \\
            \color{black}\xmark                        & \color{black}\cmark & 67.25 \\
            \color{black}\cmark                    & \color{black}\xmark     & 68.49 \\
            \rowcolor{gray!25} \color{black}\cmark & \color{black}\cmark & \textbf{68.52} \\
            \bottomrule
        \end{tabularx}
    \end{subtable}
\end{table*}

\textbf{Role of System Components.}~In Table~\ref{tab:mentor-role}, we observe a significant improvement when moving from single-teacher distillation (row 2) to a multi-mentor setup (row 3). The presence of multiple mentors, specifically the intermediate-sized peers, bridges the capacity gap between the large teacher and the small student. This gap is a well-known limitation in traditional KD, where the student struggles to fully comprehend the knowledge transferred from a much larger teacher. Introducing peers, which have capacities between the teacher and student, effectively provides a smoother learning gradient for the student, facilitating a more gradual and interpretable knowledge transfer.

The \textbf{adaptive distillation strategy} (row 4) boosts accuracy by 2.56\%, highlighting the limitations of static distillation methods. By adjusting distillation based on the student’s progress and mentor outputs, ClassroomKD ensures more efficient learning, especially during critical phases where mentor usefulness varies.  Table~\ref{tab:classroom-modules} shows that the KF Module improves accuracy from 65.96\% to 68.49\% by filtering out irrelevant knowledge, while the Mentoring Module dynamically adapts teaching strategies, raising performance to 67.25\%. Together, these modules achieve the highest accuracy of 68.52\%, ensuring both quality and adaptability in knowledge transfer.

We examine the classroom composition and further analyze our framework in the following sections.

\subsection{Classroom Size and Composition}

This section examines the impact of both the number and diversity of mentors on student performance within ClassroomKD. Our experiments investigate different mentor configurations, including varying mentor quantities and diverse architectures and performance levels.

\begin{figure*}[ht]
    \centering
    \begin{subfigure}[b]{0.45\textwidth}
        \centering
        \scriptsize
        \hspace{-0.3cm}
        \begin{tikzpicture}
            \begin{axis}[
                scale only axis,
                width=0.8\linewidth,
                height=5cm,
                title={S: MBV2, T: R50, 5 Diverse Peers},
                xlabel={Training Epoch},
                ylabel={Validation Accuracy (Top-1 \%)},
                xtick={0,50,100,150,200},
                ytick={20,30,40,50,60},
                xlabel near ticks,
                ylabel near ticks,
                xmin=0, xmax=240, 
                ymin=10, ymax=70,
                grid=major,
                grid style={dotted,color=black},
                label style={font=\scriptsize},
                tick label style={font=\scriptsize},
                title style={yshift=-4pt},
            ]

            \addplot graphics [
                xmin=0, xmax=243,
                ymin=10, ymax=70
            ] {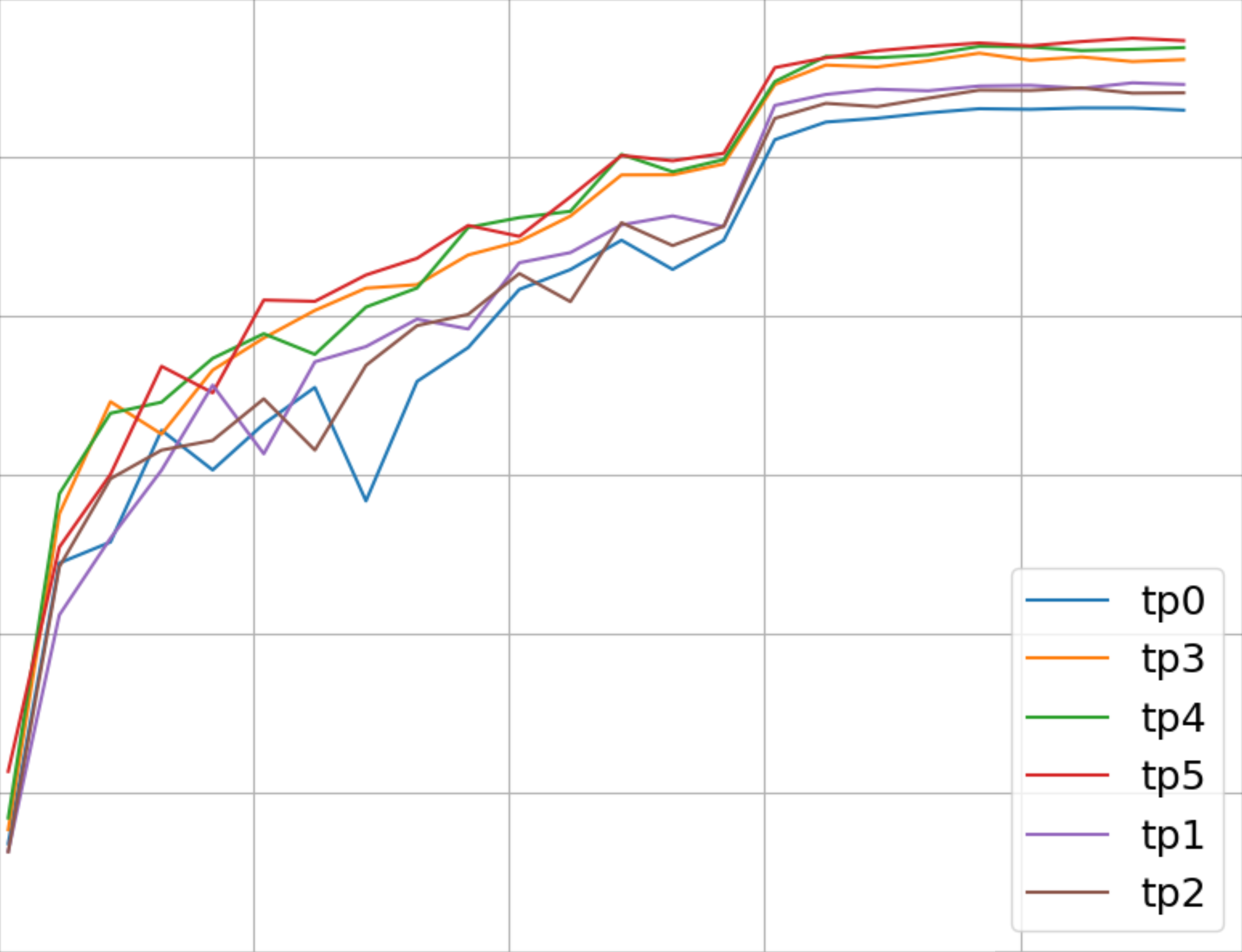};

            \end{axis}
        \end{tikzpicture}
        \caption{Without controlling mentor architectures or performance levels, we train a student in classrooms with up to six mentors. The validation accuracy improves as the number of peers increases, but the marginal gain diminishes beyond five peers. We use these results to limit the size of our classrooms to six mentors in all subsequent experiments.}
        \label{fig:classroom_size}
    \end{subfigure}
    \hfill
    \begin{subfigure}[b]{0.53\textwidth}
        \centering
        \scriptsize
        \begin{tikzpicture}
            \begin{axis}[
                scale only axis,
                width=0.85\linewidth,
                height=1.5cm,
                title={S: MBV2, T: EN-B0, 5 Same Peers},
                xlabel={Number of Mentors},
                ylabel={Student Top-1},
                xlabel near ticks,
                ylabel near ticks,
                ymin=63, ymax=68.5,
                xtick={1,2,3,4,5,6},
                grid=major,
                grid style={dotted,color=black},
                label style={font=\scriptsize},
                tick label style={font=\scriptsize},
                title style={yshift=-4pt},
            ]
        
            \addplot[
                color=purple,
                mark=o,
                thick
            ] coordinates {
                (1, 63.35)
                (2, 64.71)
                (3, 64.72)
                (4, 66.57)
                (5, 67.08)
                (6, 67.53)
            };
        
            \node at (axis cs: 1, 63.35) [anchor=south] {63.35};
            \node at (axis cs: 2, 64.71) [anchor=south] {64.71};
            \node at (axis cs: 3, 64.72) [anchor=south] {64.72};
            \node at (axis cs: 4, 66.57) [anchor=north] {66.57};
            \node at (axis cs: 5, 67.08) [anchor=north] {67.08};
            \node at (axis cs: 6, 67.53) [anchor=north] {67.53};
        
            \end{axis}
        \end{tikzpicture}
        \caption{Fixing mentor architecture and size by using multiple instances of the same mentor at different training checkpoints, we observe that student accuracy still improves with the number of mentors. This indicates that diversity in mentor performance alone is enough to enhance student learning.}
        \label{fig:peers_quantity}

        \begin{tikzpicture}
            \begin{axis}[
                scale only axis,
                width=0.85\linewidth,
                height=1.3cm,
                title={S: MBV2, T: R50, 5 Diverse Peers},
                xlabel={Number of Mentors},
                ylabel={Student Top-1},
                xlabel near ticks,
                ylabel near ticks,
                grid=major,
                grid style={dotted,color=black},
                label style={font=\scriptsize},
                legend style={draw=none,font=\scriptsize},
                legend pos=north west,
                tick label style={font=\scriptsize},
                title style={yshift=-4pt},
            ]
        
            \addplot[
                color=orange,
                mark=x,
                thick
            ] coordinates {
                (4, 67.88)
                (5, 68.33)
                (6, 68.93)
            };
            \addlegendentry{AVER}
        
            \addplot[
                color=purple,
                mark=o,
                thick
            ] coordinates {
                (4, 68.06)
                (5, 69.11)
                (6, 69.78)
            };
            \addlegendentry{Ours}
        
            \end{axis}
        \end{tikzpicture}
        \caption{Comparing our approach to vanilla multi-mentor distillation (AVER) highlights the benefit of our adaptive distillation with dynamic mentor selection as the classroom grows.}
    \end{subfigure}
    \caption{\textbf{Effect of Classroom Size and Composition.} We investigate the effect of mentor count, their architectures, and performance differences on learning.}
\end{figure*}

\textbf{Impact of peer quantity.}~Figure~\ref{fig:classroom_size} and~\ref{fig:peers_quantity} illustrates the effect of increasing the number of peers in the classroom. Without any peers, the student achieves 63.35\% top-1 accuracy. However, as peers are added, performance steadily improves, reaching 67.53\% with five peers. This improvement demonstrates that incorporating intermediate mentors (peers) with varied capacities helps bridge the gap between the large teacher and small student, making knowledge transfer more effective. However, the performance improvement plateaus beyond five peers. This suggests that while adding mentors benefits learning, the gain diminishes beyond a certain point due to redundancy in the knowledge being transferred. Therefore, we limit our classrooms to six mentors in all subsequent experiments to balance efficiency and performance.

\begin{table*}[ht]
    \centering
    \caption{\textbf{Effect of Mentor Diversity.} We investigate the role of mentor diversity in terms of architecture and performance levels.}
    \begin{subtable}[t]{0.41\textwidth}
        \centering
        \scriptsize
        \caption{\textbf{Diversity in mentor architectures.} Diverse mentor architectures improve distillation performance compared to a homogeneous setup, even when the parameter count of the diverse mentors is lower. This indicates that architectural diversity provides valuable learning signals.}
        \label{tab:mentor_diversity_arch}
        \begin{tabularx}{\linewidth}{lXcc}
            \toprule
            Classroom & Mentors & Params & Top-1 \\
            \cmidrule(lr){1-1}
            \cmidrule(lr){2-2}
            \cmidrule(lr){3-3}
            \cmidrule(lr){4-4}
            Same & EN-B0 x6 & 24.8M & 67.53 \\
            \rowcolor{gray!25}
            Diverse & VGG13, R8, R14, R20, SV1, SV2 & \textbf{12.3M} & \textbf{68.52} \\
            \bottomrule
        \end{tabularx}

    \end{subtable}
    \hfill
    \begin{subtable}[t]{0.54\textwidth}
        \centering
        \scriptsize
        \caption{\textbf{Diversity in mentor performance.} Classrooms with low-performing mentors, average mentors (a mix of medium and high performers), and diverse mentors (a combination of low, medium, and high performers) are compared. The diverse group, with a balanced mix of performance levels, yields the best student accuracy, highlighting the benefit of including mentors with varied accuracy for effective distillation.}
        \label{tab:mentor_diversity_performance}
        \begin{tabularx}{\linewidth}{Xcccc}
            \toprule
            Mentors                       & 20-50\%                          & 50-65\%                          & 65-73\%              & Top-1 \\
            \cmidrule(lr){1-1}
            \cmidrule(lr){2-2}
            \cmidrule(lr){3-3}
            \cmidrule(lr){4-4}
            \cmidrule(lr){5-5}
            Low                        & \cmark \cmark \cmark & \cmark\cmark             & -                    & 67.77   \\
            Average                    & -                                & \cmark \cmark \cmark & \cmark\cmark & 67.53   \\
            \rowcolor{gray!25}
            Diverse & \cmark                       & \cmark \cmark            & \cmark\cmark & \textbf{68.29} \\       
            \bottomrule
        \end{tabularx}
    \end{subtable}
\end{table*}

\textbf{Architectural Diversity} (Table~\ref{tab:mentor_diversity_arch}): We observe that using mentors with diverse architectures (e.g., VGG, ResNet, and ShuffleNet) yields better performance (68.52\%) compared to using multiple instances of the same architecture (67.53\%). Interestingly, this improvement occurs despite the fact that the total parameter count of the diverse mentors (12.3M) is significantly lower than that of the homogeneous set (24.8M). This indicates that architectural diversity introduces richer and more varied learning signals, which are more effective for knowledge distillation.

\textbf{Performance Diversity} (Table~\ref{tab:mentor_diversity_performance}): We also evaluate the effect of mentor performance diversity by creating classrooms composed of mentors from different performance brackets. When mentors are homogeneous in terms of performance (either all low- or all high-performing), student performance remains lower. However, a diverse set of mentors, comprising both low- and high-performing peers, leads to the highest student accuracy (68.29\%). This suggests that having varied knowledge sources across performance levels provides complementary learning experiences for the student, facilitating more robust distillation.


\subsection{Temperature in Mentoring Module}

\begin{figure}[ht]
    \begin{minipage}{0.50\linewidth}
        \centering
        \begin{tikzpicture}
            \begin{axis}[
                width=1.\linewidth,
                height=4cm,
                xlabel={Training Epoch},
                ylabel={Temperature ($\tau$)},
                xlabel near ticks,
                ylabel near ticks,
                xmin=0, xmax=250, 
                ymin=0, ymax=12,
                axis on top,
                grid=major,
                grid style={dotted,color=black},
                label style={font=\scriptsize},
                legend cell align={left},
                legend style={font=\scriptsize},
                tick label style={font=\scriptsize},
            ]

            \addplot [
                color=red,
                thick
            ] coordinates {
                (0,0) (0,0) 
            };
            \addlegendentry{Teacher}
        
            \addplot [
                color=darkgreen,
                thick
            ] coordinates {
                (0,0) (0,0) 
            };
            \addlegendentry{Peers}
            
            \addplot graphics [
                xmin=0, xmax=250,
                ymin=0, ymax=12
            ] {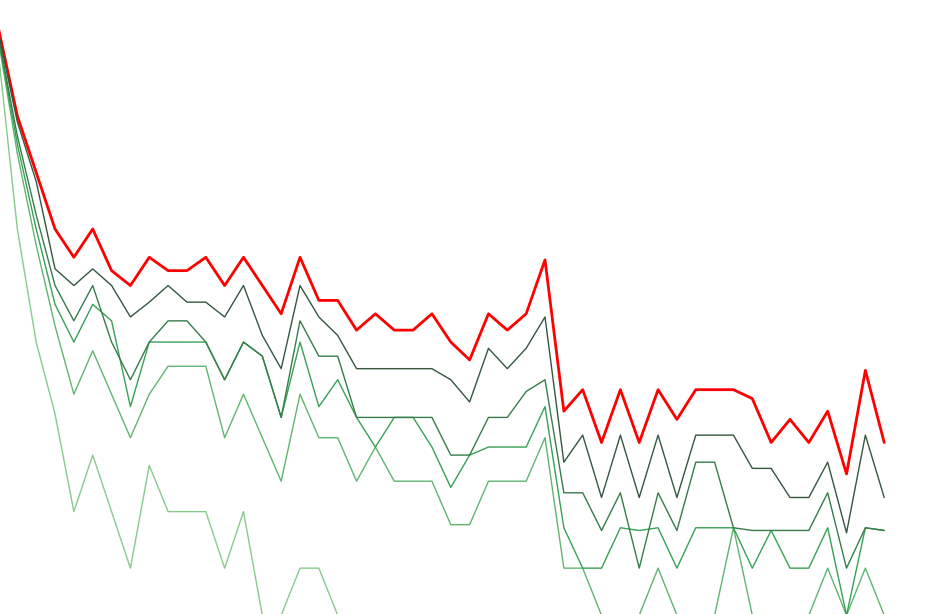};

            \end{axis}
        \end{tikzpicture}
        \vspace{-4pt}
        \caption{\textbf{Effect of temperature adaption.} Our adaptive approach independently adjusts the temperature for each mentor (teacher and peers) over time, allowing them to optimize their teaching strategies dynamically across epochs.}
        \label{fig:adaptive_temp}
    \end{minipage}
    \hfill
    \begin{minipage}{0.48\linewidth}
        \centering
        \scriptsize
        \captionof{table}{\textbf{Temperature adaption strategy.} We compare our temperature adaptation method to DTKD~\cite{wei2024dynamic} by replacing our mentoring module with their dynamic temperature computation. Our mentoring module outperforms DTKD's temperature adaption strategy with $\tau=12$ (our default) and $\tau=4$ (tuned for DTKD).}
        \label{tab:dtkd_tau}
        \begin{tabularx}{\linewidth}{XXccc}
            \toprule
            Method & Adaption & $\tau$ & MBV2 & R20 \\
            \midrule
            DTKD & DTKD & 4  & \underline{69.10} & \underline{72.05} \\
            Ours & DTKD & 4  & 64.36 & 71.18 \\
            Ours & DTKD & 12 & 68.03 & 70.02 \\
            \rowcolor{gray!25}
            Ours & Ours & 12 & \textbf{70.23} & \textbf{72.13} \\
            \bottomrule
        \end{tabularx}
    \end{minipage}
\end{figure}

We explore the role of adaptive temperature (
$\tau$) in the Mentoring Module and its impact on bridging the capacity gap between classroom networks. Our approach adjusts the temperature dynamically based on the student’s learning progress, with higher $\tau$ values at the start to accommodate the larger capacity gap, which gradually decreases as the student’s understanding improves (see fig.\ref{fig:adaptive_temp}). This adaptive strategy allows mentors to effectively ``slow down'' the teaching process during early stages and accelerate it later, ensuring effective knowledge transfer.

In our experiments, using an adaptive $\tau$ strategy yields a significant improvement in student performance. The adaptive method, which adjusts $\tau$ based on the student's progress, achieves a top-1 accuracy of 69.78\%, compared to a static $\tau$ setup where performance remains lower (65.43\% to 65.87\% for fixed values). This demonstrates that adapting the teaching pace based on the student's understanding leads to better learning outcomes.

\textbf{Comparison with DTKD.} We compared our approach with DTKD's dynamic temperature strategy by adding their method to our mentoring module. While DTKD works well with a single teacher (row 1), it is not as effective when used with multiple mentors of different capabilities. This is because DTKD assumes that all mentors predict the correct label and does not fully address the dynamic capacity gap between the teacher and student during the training process. In contrast, our method masks mentor logits with ground-truth labels, and adapts more effectively to evolving capacity gaps, achieving consistently better results across different network architectures.

\subsection{Ranking Strategies in KF Module}

\begin{figure*}[ht]
\begin{minipage}{0.48\linewidth}
    \centering
    \begin{tikzpicture}
        \begin{axis}[
            scale only axis,
            width=0.8\linewidth,
            height=3cm,
            xlabel={Training Epoch},
            ylabel={Network Rank},
            xtick={0,50,100,150,200},
            xlabel near ticks,
            ylabel near ticks,
            xmin=0, xmax=250, 
            ymin=0, ymax=1.4,
            axis on top,
            label style={font=\scriptsize},
            legend style={font=\scriptsize},
            legend columns=2,
            legend cell align={left},
            tick label style={font=\scriptsize},
            legend pos=south east,
        ]

        \addplot [
            color=red,
            thick
        ] coordinates {
            (0,0) (0,0)
        };
        \addlegendentry{Teacher}
    
        \addplot [
            color=darkgreen,
            thick
        ] coordinates {
            (0,0) (0,0)
        };
        \addlegendentry{Peers}
    
        \addplot [
            color=blue,
            thick
        ] coordinates {
            (0,0) (0,0)
        };
        \addlegendentry{Student}
    
        \addplot [
            color=black,
            thick
        ] coordinates {
            (0,0) (0,0)
        };
        \addlegendentry{Inactive}

        \addplot graphics [
            xmin=0, xmax=250,
            ymin=0, ymax=1.4
        ] {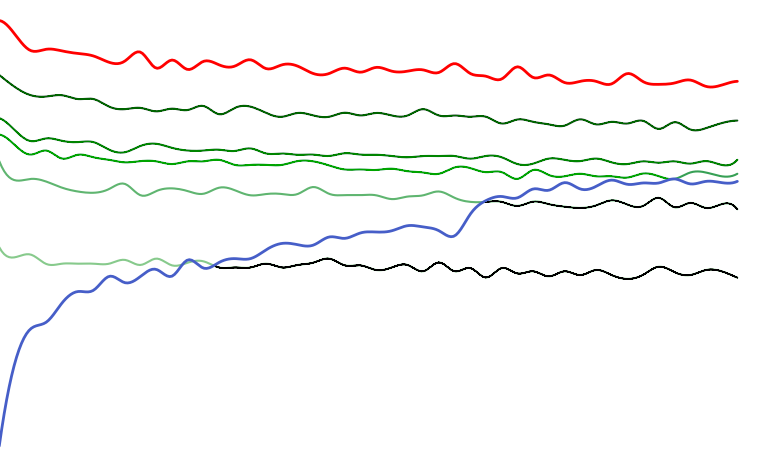};

        \end{axis}
    \end{tikzpicture}
    \vspace{-4pt}
    \caption{\textbf{Rank-based mentor activation.} Ranks evolve during training, reflecting the dynamic nature of capacity gaps. ClassroomKD uses high-quality mentors (red and green), deactivating ineffective mentors (black) who rank below the student (blue).}
    \label{fig:mentor_selection}
\end{minipage}
\hfill
\begin{minipage}{0.5\linewidth}
    \centering
    \scriptsize
    \captionof{table}{\textbf{Choice of Ranking Strategy.} We compare two ranking methods. Here, we employ different networks for peers. (A) we employ class ranks as $\alpha$, $\beta$ (B) We use discretised probabilities $\lambda$. We observe that using ranks as loss weights improves student network performance compared to probabilities.}
    \label{tab:ranking_tab}
    \begin{tabular}{lcccc}
        \toprule
        Teacher                                            & R110           & R110           & R56            & VGG13 \\
        Student                                           & R20            & R32            & R20            & VGG8 \\
        \cmidrule(lr){1-1} \cmidrule(lr){2-5}
        Method B & 71.94 & 74.28 & \textbf{72.56} & 73.58 \\
        \rowcolor{gray!25}
        Ours(A)  & \textbf{72.06} & \textbf{74.71} & 72.13 & \textbf{75.29} \\
        \midrule
        Teacher                                     & VGG13          & R32×4         & R32×4         & R50                   \\
        Student                                     & MBV2          & SN-V2         & SN-V1         & MBV2             \\
        \cmidrule(lr){1-1} \cmidrule(lr){2-5}
        Method B & 68.52 & 75.71 & 75.08 & 69.78  \\
        \rowcolor{gray!25}
        Ours(A) & \textbf{70.26} & \textbf{76.74} & \textbf{75.81} & \textbf{70.23}  \\
        \bottomrule
    \end{tabular}
\end{minipage}
\end{figure*}

We study the effect of our ranking strategy in the KF Module, which dynamically activates the teacher and peers to guide the student. In Figure~\ref{fig:mentor_selection}, we observe the evolution of ranks over time, where the teacher (red) consistently holds a higher rank than all other mentors because of its superior performance. Peer ranks (green) fluctuate, and ineffective peers are deactivated as their ranks fall below the student's rank (blue) during training. This dynamic mentor activation prevents error accumulation from underperforming mentors and allows the student to progressively improve.

In Table~\ref{tab:ranking_tab}, we explore an alternative ranking strategy (\textbf{Method B}) by replacing Eq.~\ref{eq:rank_score} with:
\begin{align}
    \bm{j} &= \text{argsort}({w^m \mid m \in \sC}) \quad \text{for } m \in \sC \\
    r^m &= \lambda \cdot \bm{j}^{-1}(m)
\end{align}
where \(r^m\) is a ranking score, \(\lambda\) is a scaling parameter set to 0.1, and \(\bm{j}^{-1}(m)\) gives the index of model \(m\) in a sorted list of weights. This results in uniformly distributed ranks (0.1, 0.2, 0.3, ...) instead of the weighted rank distribution in our original formulation. The results show that the proposed ranking method works better. However, we note that even this alternative ranking computation performs better than baseline methods for multiple networks. This improvement stems from the rank-based weighting mechanism, which focuses the student's learning on more challenging and discriminative classes, reducing sensitivity to noise and enhancing overall learning efficiency. 

\section{Conclusion}
\label{sec:conclusion}

We presented \textit{ClassroomKD}, a novel knowledge distillation framework that mimics a classroom environment, where a student learns from a diverse set of mentors. By selectively integrating feedback through the Knowledge Filtering (KF) Module and dynamically adjusting teaching strategies with the Mentoring Module, ClassroomKD ensures effective knowledge transfer and mitigates the issues of error accumulation and capacity gap. Our approach significantly improves the student’s performance in classification and pose estimation tasks, consistently outperforming traditional distillation methods.

In large-scale or real-time settings—e.g., mobile deployment for pose estimation—ClassroomKD provides a straightforward mechanism to harness existing high-capacity mentors alongside intermediate peers. As we see in the classification tasks, purely aggregating mentor logits (AVER) or using random dropping is suboptimal. Instead, by dynamically ranking mentors and adjusting teaching temperature per sample and per epoch, we reduce wasted capacity and error buildup.

The \textbf{main takeaway} is that \textit{how} knowledge is delivered—who is allowed to teach and how swiftly that teaching is introduced—can be just as crucial as which networks are present in the classroom.

\paragraph{Impact.} By fostering more efficient and adaptive students, ClassroomKD paves the way for greener AI solutions with reduced computational costs and energy consumption.
Beyond practical applications, this work encourages further research at the intersection of cognitive science and AI, enabling the exploration of more social and educational learning strategies in machine learning. For a discussion of future directions—including applying ClassroomKD to dataset distillation to further reduce memory footprint—please see Appendix~\ref{sec:future_work}.

\paragraph{Limitations.} While we demonstrated the efficacy of ClassroomKD on image classification and human pose estimation, its application to other domains and more complex tasks, such as object detection and segmentation, presents a promising avenue for future work. Despite the improvements, the framework introduces complexity, especially with respect to the mentor ranking and teaching adjustments, which can require careful tuning. Future work will explore further optimizations and expand the framework's utility to broader tasks.

\section*{Acknowledgement}

This study received partial funding from the European Union's Horizon Europe research and innovation program under grant agreement No. 101135724 (Language Augmentation for Humanverse [LUMINOUS]), addressing Topic HORIZON-CL4-2023-HUMAN-01-21.


%
%
\bibliographystyle{unsrt}
\bibliography{main}

\begin{thebibliography}{10}

\bibitem{hinton2015distilling}
Geoffrey Hinton, Oriol Vinyals, and Jeffrey Dean.
\newblock Distilling the knowledge in a neural network.
\newblock In {\em NIPS Deep Learning and Representation Learning Workshop}, 2015.

\bibitem{zagoruyko2016paying}
Sergey Zagoruyko and Nikos Komodakis.
\newblock Paying more attention to attention: Improving the performance of convolutional neural networks via attention transfer.
\newblock {\em arXiv}, abs/1612.03928, 2016.

\bibitem{adriana2015fitnets}
Romero Adriana, Ballas Nicolas, K~Samira Ebrahimi, Chassang Antoine, Gatta Carlo, and Bengio Yoshua.
\newblock Fitnets: Hints for thin deep nets.
\newblock In {\em Proceedings of the International Conference on Learning Representations}, 2015.

\bibitem{you2017learning}
Shan You, Chang Xu, Chao Xu, and Dacheng Tao.
\newblock Learning from multiple teacher networks.
\newblock In {\em Proceedings of the 23rd ACM SIGKDD international conference on knowledge discovery and data mining}, pages 1285--1294, 2017.

\bibitem{mirzadeh2019improved}
Seyed~Iman Mirzadeh, Mehrdad Farajtabar, Ang Li, Nir Levine, Akihiro Matsukawa, and Hassan Ghasemzadeh.
\newblock Improved knowledge distillation via teacher assistant.
\newblock In {\em AAAI Conference on Artificial Intelligence}, 2019.

\bibitem{son2021densely}
Wonchul Son, Jaemin Na, Junyong Choi, and Wonjun Hwang.
\newblock Densely guided knowledge distillation using multiple teacher assistants.
\newblock In {\em Proceedings of the IEEE/CVF Conference on Computer Vision and Pattern Recognition}, pages 9395--9404, 2021.

\bibitem{hao2024one}
Zhiwei Hao, Jianyuan Guo, Kai Han, Yehui Tang, Han Hu, Yunhe Wang, and Chang Xu.
\newblock One-for-all: Bridge the gap between heterogeneous architectures in knowledge distillation.
\newblock {\em Advances in Neural Information Processing Systems}, 36, 2024.

\bibitem{passalis2018learning}
Nikolaos Passalis and Anastasios Tefas.
\newblock Learning deep representations with probabilistic knowledge transfer.
\newblock In {\em Proceedings of the European Conference on Computer Vision (ECCV)}, pages 268--284, 2018.

\bibitem{kim2018paraphrasing}
Jangho Kim, SeongUk Park, and Nojun Kwak.
\newblock Paraphrasing complex network: Network compression via factor transfer.
\newblock {\em Advances in neural information processing systems}, 31, 2018.

\bibitem{heo2019knowledge}
Byeongho Heo, Minsik Lee, Sangdoo Yun, and Jin~Young Choi.
\newblock Knowledge transfer via distillation of activation boundaries formed by hidden neurons.
\newblock In {\em Proceedings of the AAAI conference on artificial intelligence}, volume~33, pages 3779--3787, 2019.

\bibitem{ahn2019variational}
Sungsoo Ahn, Shell~Xu Hu, Andreas Damianou, Neil~D Lawrence, and Zhenwen Dai.
\newblock Variational information distillation for knowledge transfer.
\newblock In {\em Proceedings of the IEEE/CVF Conference on Computer Vision and Pattern Recognition}, pages 9163--9171, 2019.

\bibitem{tian2020contrastive}
Yonglong Tian, Dilip Krishnan, and Phillip Isola.
\newblock Contrastive representation distillation.
\newblock In {\em Proceedings of the International Conference on Learning Representations}, 2020.

\bibitem{park2019relational}
Wonpyo Park, Dongju Kim, Yan Lu, and Minsu Cho.
\newblock Relational knowledge distillation.
\newblock In {\em Proceedings of the IEEE/CVF Conference on Computer Vision and Pattern Recognition}, pages 3967--3976, 2019.

\bibitem{tung2019similarity}
Frederick Tung and Greg Mori.
\newblock Similarity-preserving knowledge distillation.
\newblock In {\em Proceedings of the IEEE/CVF Conference on Computer Vision and Pattern Recognition}, pages 1365--1374, 2019.

\bibitem{yang2021knowledge}
Jing Yang, Brais Mart{\'i}nez, Adrian Bulat, and Georgios Tzimiropoulos.
\newblock Knowledge distillation via softmax regression representation learning.
\newblock In {\em Proceedings of the International Conference on Learning Representations}, 2021.

\bibitem{huang2022knowledge}
Tao Huang, Shan You, Fei Wang, Chen Qian, and Chang Xu.
\newblock Knowledge distillation from a stronger teacher.
\newblock {\em Advances in Neural Information Processing Systems}, 35:33716--33727, 2022.

\bibitem{zhou2021rethinking}
Helong Zhou, Liangchen Song, Jiajie Chen, Ye~Zhou, Guoli Wang, Junsong Yuan, and Qian Zhang.
\newblock Rethinking soft labels for knowledge distillation: A bias-variance tradeoff perspective.
\newblock In {\em Proceedings of the International Conference on Learning Representations}, 2021.

\bibitem{lin2022knowledge}
Sihao Lin, Hongwei Xie, Bing Wang, Kaicheng Yu, Xiaojun Chang, Xiaodan Liang, and Gang Wang.
\newblock Knowledge distillation via the target-aware transformer.
\newblock In {\em Proceedings of the IEEE/CVF Conference on Computer Vision and Pattern Recognition}, pages 10915--10924, 2022.

\bibitem{li2023curriculum}
Zheng Li, Xiang Li, Lingfeng Yang, Borui Zhao, Renjie Song, Lei Luo, Jun Li, and Jian Yang.
\newblock Curriculum temperature for knowledge distillation.
\newblock In {\em Proceedings of the AAAI Conference on Artificial Intelligence}, volume~37, pages 1504--1512, 2023.

\bibitem{wei2024dynamic}
Yukang Wei and Yu~Bai.
\newblock Dynamic temperature knowledge distillation.
\newblock {\em arXiv preprint arXiv:2404.12711}, 2024.

\bibitem{zhang2018mutual}
Ying Zhang, Tao Xiang, Timothy~M. Hospedales, and Huchuan Lu.
\newblock Deep mutual learning.
\newblock In {\em Proceedings of the IEEE/CVF Conference on Computer Vision and Pattern Recognition}, June 2018.

\bibitem{zhu2018knowledge}
Xiatian Zhu, Shaogang Gong, et~al.
\newblock Knowledge distillation by on-the-fly native ensemble.
\newblock {\em Advances in neural information processing systems}, 31, 2018.

\bibitem{chen2020online}
Defang Chen, Jian-Ping Mei, Can Wang, Yan Feng, and Chun Chen.
\newblock Online knowledge distillation with diverse peers.
\newblock In {\em AAAI}, pages 3430--3437, 2020.

\bibitem{li2020online}
Zheng Li, Ying Huang, Defang Chen, Tianren Luo, Ning Cai, and Zhigeng Pan.
\newblock Online knowledge distillation via multi-branch diversity enhancement.
\newblock In {\em Proceedings of the Asian Conference on Computer Vision (ACCV)}, November 2020.

\bibitem{kd_pose7}
Zheng Li, Jingwen Ye, Mingli Song, Ying Huang, and Zhigeng Pan.
\newblock Online knowledge distillation for efficient pose estimation.
\newblock In {\em Proceedings of the IEEE/CVF Conference on Computer Vision and Pattern Recognition}, pages 11740--11750, 2021.

\bibitem{li2022shadow}
Lujun Li and Zhe Jin.
\newblock Shadow knowledge distillation: Bridging offline and online knowledge transfer.
\newblock {\em Advances in Neural Information Processing Systems}, 35:635--649, 2022.

\bibitem{du2020agree}
Shangchen Du, Shan You, Xiaojie Li, Jianlong Wu, Fei Wang, Chen Qian, and Changshui Zhang.
\newblock Agree to disagree: Adaptive ensemble knowledge distillation in gradient space.
\newblock In H.~Larochelle, M.~Ranzato, R.~Hadsell, M.F. Balcan, and H.~Lin, editors, {\em Advances in Neural Information Processing Systems}, volume~33, pages 12345--12355. Curran Associates, Inc., 2020.

\bibitem{krizhevsky2009learning}
Alex Krizhevsky, Geoffrey Hinton, et~al.
\newblock Learning multiple layers of features from tiny images.
\newblock {\em Master's thesis, Department of Computer Science, University of Toronto}, 2009.

\bibitem{deng2009imagenet}
Jia Deng, Wei Dong, Richard Socher, Li-Jia Li, Kai Li, and Li~Fei-Fei.
\newblock Imagenet: A large-scale hierarchical image database.
\newblock In {\em 2009 IEEE conference on computer vision and pattern recognition}, pages 248--255. Ieee, 2009.

\bibitem{lin2014microsoft}
Tsung-Yi Lin, Michael Maire, Serge Belongie, James Hays, Pietro Perona, Deva Ramanan, Piotr Doll{\'a}r, and C~Lawrence Zitnick.
\newblock Microsoft coco: Common objects in context.
\newblock In {\em Computer Vision--ECCV 2014: 13th European Conference, Zurich, Switzerland, September 6-12, 2014, Proceedings, Part V 13}, pages 740--755. Springer, 2014.

\bibitem{andriluka20142d}
Mykhaylo Andriluka, Leonid Pishchulin, Peter Gehler, and Bernt Schiele.
\newblock 2d human pose estimation: New benchmark and state of the art analysis.
\newblock In {\em Proceedings of the IEEE Conference on computer Vision and Pattern Recognition}, pages 3686--3693, 2014.

\bibitem{chen2022knowledge}
Defang Chen, Jian-Ping Mei, Hailin Zhang, Can Wang, Yan Feng, and Chun Chen.
\newblock Knowledge distillation with the reused teacher classifier.
\newblock In {\em Proceedings of the IEEE/CVF Conference on Computer Vision and Pattern Recognition}, pages 11933--11942, 2022.

\bibitem{zhang2022confidence}
Hailin Zhang, Defang Chen, and Can Wang.
\newblock Confidence-aware multi-teacher knowledge distillation.
\newblock In {\em ICASSP 2022-2022 IEEE International Conference on Acoustics, Speech and Signal Processing (ICASSP)}, pages 4498--4502. IEEE, 2022.

\bibitem{DBLP:journals/corr/abs-2107-03332}
Yanjie Li, Sen Yang, Shoukui Zhang, Zhicheng Wang, Wankou Yang, Shu{-}Tao Xia, and Erjin Zhou.
\newblock Is 2d heatmap representation even necessary for human pose estimation?
\newblock {\em CoRR}, abs/2107.03332, 2021.

\bibitem{zhou2022dataset}
Yongchao Zhou, Ehsan Nezhadarya, and Jimmy Ba.
\newblock Dataset distillation using neural feature regression.
\newblock In {\em Proceedings of the Advances in Neural Information Processing Systems (NeurIPS)}, 2022.

\end{thebibliography}


\clearpage
\appendix

\section{Analysis and Additional Results}

\subsection{Per-Class Performance Improvement}

We further analyze ClassroomKD's effectiveness by examining the per-class performance improvements of the distilled student model compared to the baseline model (without knowledge distillation). To this end, we compare the class-level accuracy differences between ClassroomKD and a standard multi-teacher knowledge distillation (AVER) approach, both using the distilled CIFAR-100 dataset.

\begin{figure}[ht]
    \centering
        \centering
        \begin{tikzpicture}
            \begin{axis}[
                scale only axis,
                width=0.4\linewidth,
                height=3cm,
                xlabel={Class Index},
                ylabel={Performance Gain (\%)},
                xlabel near ticks,
                ylabel near ticks,
                xmin=0, xmax=100, 
                ymin=-12, ymax=24,
                axis on top,
                label style={font=\scriptsize},
                tick label style={font=\scriptsize},
                title style={font=\scriptsize},
                title={ClassroomKD}
            ]
    
            \addplot graphics [
                xmin=-4.5, xmax=104.5,
                ymin=-8, ymax=24
            ] {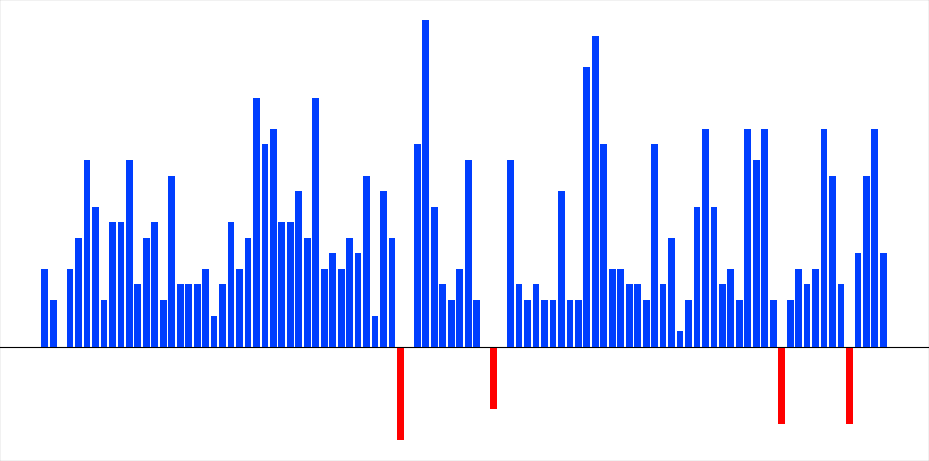};
    
            \end{axis}
        \end{tikzpicture}
        \begin{tikzpicture}
            \begin{axis}[
                scale only axis,
                width=0.4\linewidth,
                height=3cm,
                xlabel={Class Index},
                xlabel near ticks,
                xmin=0, xmax=100, 
                ymin=-12, ymax=24,
                axis on top,
                ytick=\empty,
                label style={font=\scriptsize},
                tick label style={font=\scriptsize},
                title style={font=\scriptsize},
                title={AVER}
            ]
    
            \addplot graphics [
                xmin=-4.5, xmax=104.5,
                ymin=-11, ymax=11
            ] {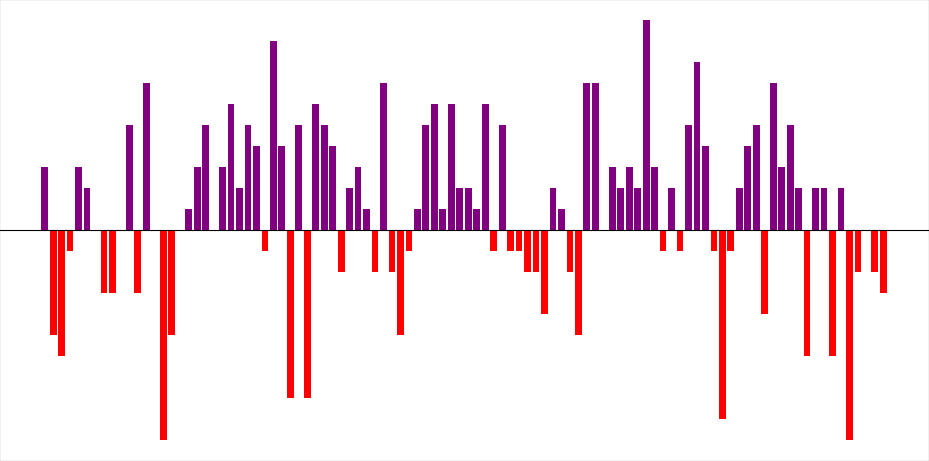};
    
            \end{axis}
        \end{tikzpicture}
    \caption{\textbf{Comparison of per-class performance gain over the NOKD baseline.} With ClassroomKD (left), the distilled model improves performance on 86 classes. With multi-teacher KD without mentor ranking (right), much fewer classes improve, the absolute improvement is smaller, and the remaining classes experience larger performance degradation (red bars). This highlights the impact of our dynamic strategies in improving performance across different classes.}
    \label{fig:per_class_performance}
\end{figure}

In Figure~\ref{fig:per_class_performance}, we illustrate the performance differences between the ClassroomKD student and the baseline model on the left. ClassroomKD improves performance in 86 out of 100 classes while minimizing performance degradation in the remaining classes. In contrast, AVER (right) has a significantly smaller improvement, and the absolute performance degradation is more severe than with ClassroomKD. This demonstrates the benefit of our mentor ranking strategy, which dynamically selects mentors based on their relative performance and reduces the likelihood of detrimental knowledge transfer or error accumulation from multiple mentors.

\subsection{Intuition Behind Proposed Ranking Method}

\begin{figure}[ht]
    \includegraphics[width=\linewidth]{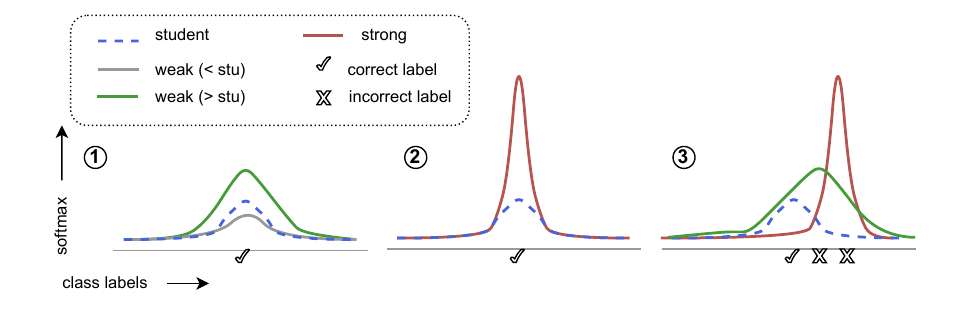}
    \caption{\textbf{Illustration of the classroom models' probabilistic distributions} The student encounters three types of mentors while learning: 1. weak classifiers predict with low confidence. 2. strong classifiers are highly confident in their prediction. 3. Wrong classifiers predict incorrect labels.}
    \label{fig:softmax_cases}
\end{figure}

\textbf{Classroom Dynamics.} For a given sample \(x_k\), we can visualize the output probability distribution of a model \(m\) by plotting the softmax probability \(P_{z_i}^m\) of its logit \(z_i\) against the class labels \(i\), for all \(i \in C\). The models in a classroom can have logit distributions that fall into one of the three cases: (1) Weak classifiers predict the true label \(y_k\) with low confidence. (2) Strong classifiers predict the true class with high confidence, giving it a "sharper" peak. (3) Wrong classifiers have a peak at the wrong class. This is illustrated in Fig.~\ref{fig:softmax_cases}. Using \(P_{z_i}^m\) based ranks will help the student learn to filter out wrong classifiers.

\subsection{Dynamic Capacity Gap Visualization}
\label{sec:dynamic-capacity-gap}

To better understand probabilistic distributions (Fig.~\ref{fig:softmax_cases}) of our classroom, we plot the softmax of the logits produced by the student model and mentors at various training steps.

\begin{figure}[ht]
    \includegraphics[width=\linewidth]{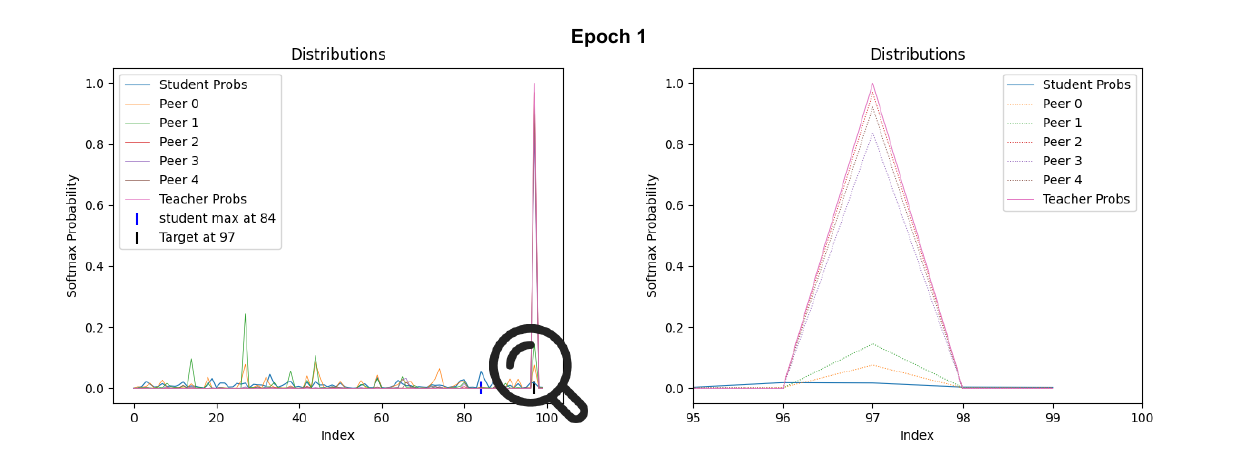}
    \caption{Probability Distributions at Epoch 1. Right subplot is zoomed in at the true class (97).}
    \label{fig:softmax_ep1}
\end{figure} 

\begin{figure}[ht]
    \includegraphics[width=\linewidth]{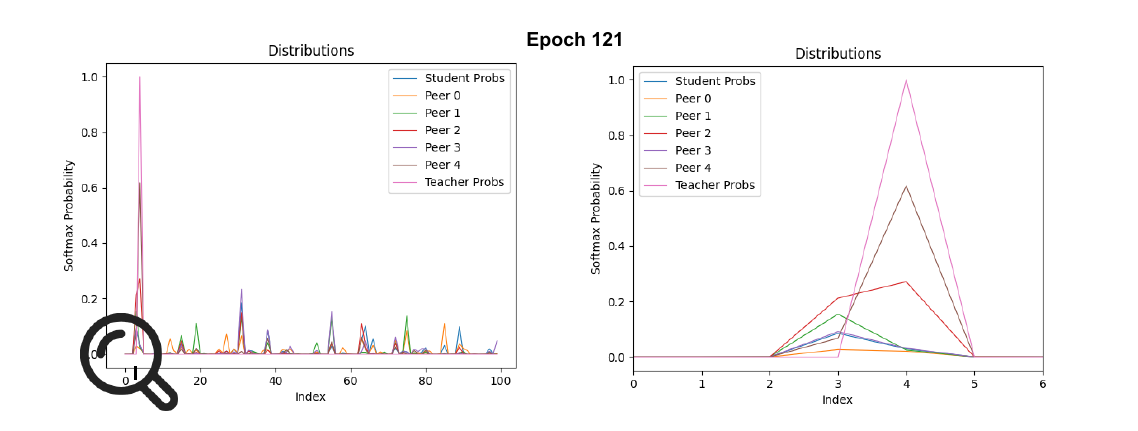}
    \caption{Probability Distributions at Epoch 121. Right subplot is zoomed in at the true class (4)}
    \label{fig:softmax_ep121}
\end{figure}

\begin{figure}[ht]
    \includegraphics[width=\linewidth]{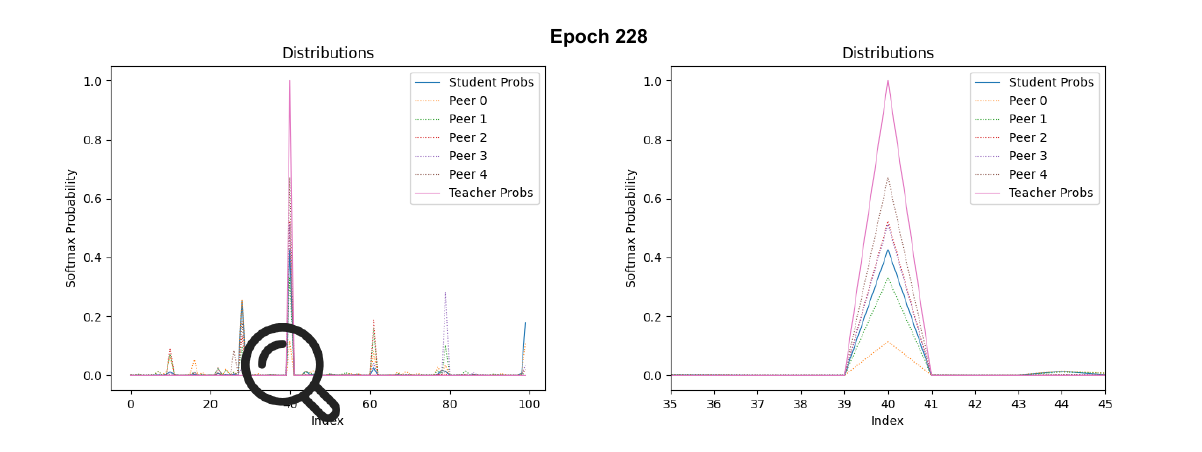}
    \caption{Probability Distributions at Epoch 228. Right subplot is zoomed in at the true class (40)}
    \label{fig:softmax_ep228}
\end{figure}

We observe a gradual decrease in the gap between the student and teacher's probabilities at the true label from epochs 1 through 228.(Fig.~\ref{fig:softmax_ep1}, ~\ref{fig:softmax_ep121} and ~\ref{fig:softmax_ep228})




\clearpage
\section{ClassroomKD for 2D Human Pose Estimation}
\label{sec:pose_estimation}

The proposed methodology can be applied to distill knowledge to smaller models in 2D HPE with a few modifications. 


\subsection{Top-down SimCC-Based Methods}

RTMPose architecture, which we use for our experiments on the COCO Kepoints dataset, contains a SimCC~\cite{DBLP:journals/corr/abs-2107-03332} head that outputs separate logits of the shape (N, K, D) each in the x and y directions, where N is the batch size, K is the number of joints, and D is the coordinate dimensions. For our purposes, only K is relevant. This output can be seen as \textbf{two predictions} for each of the K joints. Hence, we apply the following three modifications to adapt our approach:

\begin{enumerate}
\item The sharpness of model m, \(P^m\), is calculated using the PCK accuracy metric. These values are further normalized in the classroom to obtain their respective ranks.
\item Once the \textit{active} mentors are chosen, the $\mathcal{L}_{\mathrm{distill}}$ is processed as the combined distillation loss between the student and mentor along x and y directions.
\item The logits' shapes are converted to (N*K,-1) before applying the KL-divergence. The sum of distillation losses along the x and y directions is finally divided by the number of joints.
\begin{equation} \mathcal{L}_{\mathrm{simcc}}(\hat{\vy}^m,\hat{\vy}^s;\tau^m) = \frac{1}{K} (\mathcal{L}_{\mathrm{distill}}(\hat{\vy}_x^m,\hat{\vy}_x^s;\tau^m) + \mathcal{L}_{\mathrm{distill}}(\hat{\vy}_y^m,\hat{\vy}_y^s;\tau^m)) \end{equation}
\end{enumerate}

\subsection{Top-down Heatmap-Based Methods}

The LiteHRNet model, which we use for our experiments on the MPII Human Pose dataset, outputs 2D heatmaps of size (N, K, H, W). This is equivalent to the two separate 1D heatmaps in SimCC heads. To apply ClassroomKD in this case, we make the below changes:

\begin{enumerate}
\item Similar to the SimCC head, the sharpness of model m, \(P^m\), is calculated using the PCK metric for the ranking.
\item The KL-divergence between the student and \textit{active} mentors is calculated between the heatmaps and is then divided by the number of joints. 
\end{enumerate}

\clearpage
\section{ClassroomKD Algorithm}
\label{algorithm}
\begin{algorithm}[ht]
\caption{ClassroomKD}
\label{alg:weighted_loss}
\begin{algorithmic}[1]
\REQUIRE Input batch $\mathbf{x}$
\REQUIRE Ground truth labels $\mathbf{y}$
\REQUIRE Student $s$
\REQUIRE Mentors $\mathbb{M} \gets \{t\} \cup \{p_i\}_{i=1}^n$
\REQUIRE $\beta$: weight of distillation loss
\REQUIRE $\delta$: weight of standard KD loss with the teacher
\STATE $\text{weights} \gets \{\}$ \hfill // Initialize empty dictionary for mentor weights
\STATE $\text{ranks} \gets \{\}$ \hfill // Initialize empty dictionary for mentor ranks
\STATE $\mathcal{L} \gets 0$ \hfill // Initialize total loss 

\STATE $\mathbb{C} \gets \{s\} \cup \mathbb{M}$
\FOR{$m \in \mathbb{C}$}
    \STATE $\hat{\vy}^m \gets m(\vx)$ \hfill // Get predictions from model $m$
    \STATE $\vp^m_{\text{gt}} \gets 1/(\text{exp}(\mathrm{CELoss}(\hat{\vy}^m, \text{targets}))$ \hfill // Isolate probabilities assigned to ground truth
    \STATE $w^m \gets \text{average}(\vp^m_{\text{gt}}, \text{dim-1})$ \hfill // Average correct class probability for model $m$
    \STATE $\text{weights}[m] \gets w^m$ \hfill // Store weight for model $m$
\ENDFOR

\STATE $\text{weights} \gets \text{dict(sorted(weights.items(), key=lambda item: item[1]))}$ \hfill // Sort
\STATE $\text{total\_weight} \gets \sum(\text{weights.values()})$ \hfill // Calculate sum of all mentor weights
\STATE $\text{ranks} \gets \{m: (|\mathbb{M}| \cdot w)/{\text{total\_weight}} \text{ for } m, w \in \text{weights.items()} \}$ \hfill // Assign ranks 

\FOR{$m \in \mathbb{M}$}
    \IF{$\text{ranks}[m] > \text{ranks}[s]$}
        \STATE $\tau^m \gets \frac{\text{ranks}[m] - \text{ranks}[s]}{\text{ranks}[m]}$
        \STATE $\mathcal{L}_{\mathrm{distill}} \gets \mathrm{KL}(\hat{\vy}^m, \hat{\vy}^s, \tau^m)$
        \STATE $\mathcal{L}_{\mathrm{distill}} \gets \text{ranks}[m] \cdot \mathcal{L}_{\mathrm{distill}}$
    \ELSE
        \STATE $\mathcal{L}_{\mathrm{distill}} \gets 0$
    \ENDIF
    \STATE $\mathcal{L} \gets \mathcal{L} + \mathcal{L}_{\mathrm{distill}}$ \hfill // Add distillation loss to total loss
\ENDFOR

\STATE $\mathcal{L}_{\mathrm{task}} \gets \mathrm{CELoss}(\hat{\vy}^s, \text{targets})$ \hfill // Compute task loss (e.g., cross-entropy)
\STATE $\mathcal{L}_{\mathrm{classroom}} \gets \text{ranks}[s] \cdot \mathcal{L}_{\mathrm{task}} + \beta \cdot \mathcal{L}$ \hfill // Weight task loss by student's rank
\STATE $\mathcal{L}_{\mathrm{KD}}^t \gets \mathcal{L}_{\mathrm{task}} + \mathrm{KL}(\hat{\vy}^t, \hat{\vy}^s, \tau^t=1) $ \hfill // Compute standard student-teacher KD loss
\STATE $\mathcal{L} \gets \delta\mathcal{L}_{\mathrm{KD}}^t +  \beta \cdot \mathcal{L}_{\mathrm{classroom}}$ \hfill // Combine KD loss and classroom distillation

\RETURN $\mathcal{L}$ \hfill // Return the total loss
\end{algorithmic}
\end{algorithm}

\section{Training Protocols}
\label{sec:mentor-configuration}

\textbf{Mentor Configuration.} We use a predefined order for the mentor set in all experiments for consistency. Any deviations from this are clearly stated.

\begin{table}
    \centering
    \scriptsize
    \caption{\textbf{Mentor configurations} used in all our experiments, along with their respective top-1 accuracies and ensemble performance. The size of the mentors, should all the peers be replaced by the teacher ($(n+1)t$), the size of the current mentors ($1tnp$), and the student size are also included.}
    \label{tab:classroom-configuration}
    \begin{tabular}{lccccccccccc}
    \toprule
    & \multicolumn{6}{c}{Mentors} & \multicolumn{3}{c}{Params (M)} \\
    \cmidrule(lr){2-7} \cmidrule(lr){8-10}
    $s$ & $t$ & $p_{1}$ & $p_{2}$ & $p_{3}$ & $p_{4}$ & $p_{5}$ & $(n+1)t$ & $1tnp$ & $s$ \\
    \midrule
    \multicolumn{9}{c}{\textbf{CIFAR-100 Classification}} \\
    \midrule
    R20 & R110 & R8 & R14& SN-V2&MBV2&SN-V1 & \\
    (69.06) & (74.31) & (60.22) & (67.28) & (72.60) & (63.51) & (71.29) & 10.42 & 5.12 & 0.27 \\
    \cmidrule(lr){1-1} \cmidrule(lr){2-7} \cmidrule(lr){8-10}
    R32 & R110 & R8 & R14& SN-V2&MBV2&SN-V1 & \\
    (71.14) & (74.31) & (60.22) & (67.28) & (72.60) & (63.51) & (71.29) & 10.42 & 5.12 & 0.47 \\
    \cmidrule(lr){1-1} \cmidrule(lr){2-7} \cmidrule(lr){8-10}
    R20 & R56 & R8 & R14& SN-V2&MBV2&SN-V1 \\
    (69.06) & (72.41) & (60.22) & (67.28) & (72.60) & (63.51) & (71.29) & 5.17 & 4.24 & 0.27 \\
    \cmidrule(lr){1-1} \cmidrule(lr){2-7} \cmidrule(lr){8-10}
    VGG8 & VGG13 & R20 & MBV2& SN-V2&R56&R110 & \\
    (70.36) & (74.64) & (69.06) & (63.51) & (72.60) & (72.41) & (74.31) & 56.77 &14.50 & 3.96 \\
    \cmidrule(lr){1-1} \cmidrule(lr){2-7} \cmidrule(lr){8-10}
    MBV2 & VGG13 & R8 & R14& R20&SN-V1&SN-V2 \\
    (63.51) & (74.64) & (60.22) & (67.28) & (69.06) & (71.29) & (72.60) & 56.77 & 12.31 & 0.81 \\
    \cmidrule(lr){1-1} \cmidrule(lr){2-7} \cmidrule(lr){8-10}
    SN-V2 & R32x4 & R8 & R14& R20&MBV2&SN-V1 \\
    (72.60) & (79.42) & (60.22) & (67.28) & (69.06) & (63.51) & (71.29) & 44.62 & 9.739 &  1.35 \\
    \cmidrule(lr){1-1} \cmidrule(lr){2-7} \cmidrule(lr){8-10}
    SN-V1 & W-40-2 & R20 & MBV2 & SN-V2 & R56 & VGG13 \\
    (71.29) & (75.61) & (69.06) & (63.51) & (72.60) & (72.41) & (74.64) & 13.53 & 15.00 & 0.95 \\
    \cmidrule(lr){1-1} \cmidrule(lr){2-7} \cmidrule(lr){8-10}
    MBV2 & R50 & R8 & R14& R20&SN-V1&SN-V2 \\
    (63.51) & (79.34) & (60.22) & (67.28) & (69.06) & (71.29) & (72.60) & 142.23 & 26.55 & 0.81 \\
    \cmidrule(lr){1-1} \cmidrule(lr){2-7} \cmidrule(lr){8-10}
    SN-V1 & R32x4 & R8 & R14& R20&MBV2&SN-V2 \\
    (71.29) & (79.42) & (60.22) & (67.28) & (69.06) & (63.51) & (72.60) & 44.62 & 10.14 & 0.95 \\
    \cmidrule(lr){1-1} \cmidrule(lr){2-7} \cmidrule(lr){8-10}
    W-16-2 & W-40-2 & R20 & MBV2 & SN-V2 & R56 & VGG13  \\
    (73.64) & (75.61) & (69.06) & (63.51) & (72.60) & (72.41) & (74.64)& 13.53 & 14.98 & 0.70 \\
    \cmidrule(lr){1-1} \cmidrule(lr){2-7} \cmidrule(lr){8-10}
    MBV2 & ENB0 & ENB0 & ENB0 & ENB0 & ENB0 & ENB0  \\
    (63.51) & (73.21) & (60.23) & (61.03) & (63.60) & (66.87) & (72.70) & 24.81 & 24.81 & 0.81\\
    \cmidrule(lr){1-1} \cmidrule(lr){2-7} \cmidrule(lr){8-10}
    R18 & Swin-T(224) & SN-V2& W-40-2 & VGG13 & R32x4 &  - \\
    (74.01) & (88.78) & (72.60) & (75.61) & (74.64) & (79.42) & - & 137.98 & 48.10 & 11.22 \\
    \midrule
    \multicolumn{10}{c}{\textbf{ImageNet Classification}} \\
    \midrule
    R18 & R34 & MBV3-s & GN & MBV2 & RG-x400mf & - \\
    (69.75) & (73.31) & (67.66) & (69.79) & (71.88) & (72.83) & - & 109.00 & 39.90 & 11.70 \\
    \midrule
    \multicolumn{10}{c}{\textbf{COCO Keypoints Estimation}} \\
    \midrule
    RP-t & RP-l* & RP-s & RP-m & RP-l & - & - \\
    (68.2) & (76.5) & (71.6) & (74.6) & (75.8) & - & - & - & - & - \\
    \midrule
    \multicolumn{10}{c}{\textbf{MPII Human Pose Estimation}} \\
    \midrule
    LHR-18 & HR-W32D & LHR-30 & HR-W32 & HR-W48 & - & - \\
    (85.91) & (90.4) & (86.9) & (90.0) & (90.1) & - & - & - & - & - \\
    \cmidrule(lr){1-1} \cmidrule(lr){2-7} \cmidrule(lr){8-10}
    LHR-18 & HR-W32D & SN-V2 & MBV2 & R50 & - & - \\
    (85.91) & (90.4) & (82.8) & (85.4) & (88.2) & - & - & - & - & - \\
    \bottomrule
    \end{tabular}
\end{table}

\textbf{Model abbreviations}: MB: MobileNet, SN: ShuffleNet, R: ResNet, W: WRN, EN: EfficientNet, GN: GoogleNet, RP: RTMPose, HR: HRNet, LHR: LiteHRNet, RG: RegNet

\textbf{Hardware and Software Configuration.} We trained most of our CIFAR-100 experiments on a single V100-16GB GPU. The time required for an experiment ranged between 4 and 4.5 hours on average. We build our code on top of \texttt{Image Classification SOTA} repository\footnote{\url{https://github.com/hunto/image_classification_sota/}} and MMPose, and use pretrained models from these libraries as our mentors.

\section{Future Direction: ClassroomKD and Dataset Distillation}
\label{sec:future_work}

ClassroomKD shows strong potential in knowledge distillation, and one promising extension is its application in dataset distillation, which can further broaden its impact across various tasks.

\begin{figure}[ht]
    \centering
    \includegraphics[trim={0.5cm 0.7cm 1cm 0.5cm},clip,width=0.5\linewidth]{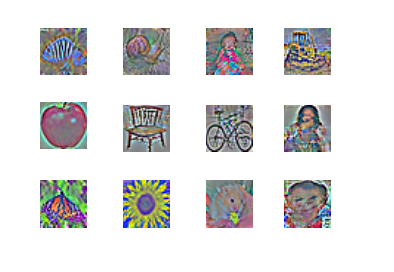}
    \caption{\textbf{Sample from the distilled CIFAR-100 dataset created using FRePo.} The dataset is reduced to 10 representative images per class, where each image encapsulates key characteristics of the class. This distilled dataset significantly reduces storage and computational requirements while maintaining essential features for effective training.}
    \label{fig:cifar100_distilled}
\end{figure}

Dataset distillation aims to create small, synthetic datasets that enable neural networks to achieve comparable performance to those trained on the original, much larger datasets. This approach reduces computational costs and storage requirements while maintaining model generalization. By optimizing a small set of representative training samples, a distilled dataset \(S\) is generated such that a model trained on \(S\) performs well on the original dataset \(\mathcal{T}\). In our experiments, we use \textbf{FRePo}~\cite{zhou2022dataset} to create a distilled CIFAR-100 dataset, reducing each class to only 10 samples (Figure~\ref{fig:cifar100_distilled}). Of these, 7 images per class are used for training, while the remaining 3 are used for testing.

As shown in Table~\ref{tab:mob_distill}, we conducted experiments on this distilled CIFAR-100 dataset and evaluated validation performance on the full CIFAR-100 dataset using the MobileNetV2 and ResNet-20 architectures. Notably, the standalone MobileNetV2 student achieves 31.00 on the distilled dataset, with 3.75\% top-1 accuracy on the full validation set. However, applying ClassroomKD with 1 teacher and 5 peers significantly improves performance, reaching 44.34 on the distilled data and 6.30\% top-1 accuracy on the full CIFAR-100 validation set. This is in stark contrast to the AVER approach, which results in only 2.33 on the distilled data and 1.51\% top-1 accuracy on the full validation set using the same number of mentors. Similarly, ClassroomKD achieves superior results with ResNet-20, showing a notable 9.66 percentage point improvement on the distilled data compared to NOKD and a 1.85 percentage point gain on the full CIFAR-100 validation set.

\begin{table}[ht]
\caption{\textbf{Performance comparison on the distilled CIFAR-100 dataset and validation metrics on the full CIFAR-100 dataset.} Results show top-1 accuracy on both the distilled dataset (7 images per class for training) and the full CIFAR-100 validation set. ClassroomKD (1 teacher, 5 peers) outperforms both the standalone student and AVER, demonstrating its efficacy in low-data regimes.}
\centering
\scriptsize
 \begin{tabular}{lccccccccccc}
  \toprule
    Student                   & \multicolumn{2}{c}{MobileNetV2} & \multicolumn{3}{c}{ResNet-20} \\ 
    \cmidrule(lr){2-3} \cmidrule(lr){4-6}
    Method                    & Distilled Top-1 & Top-1 & Distilled Top-1 & Top-1 & Top-5 \\ 
    \cmidrule(lr){1-1} \cmidrule(lr){2-2} \cmidrule(lr){3-3} \cmidrule(lr){4-4} \cmidrule(lr){5-5} \cmidrule(lr){6-6}
    NOKD        & 31.00          & 3.75          & 50.00          & 3.08          & 12.50          \\ 
    AVER        &  2.33          & 1.51          & 32.00          & 3.55          & 15.24          \\
    \rowcolor{gray!25}
    ClassroomKD & \textbf{44.34} & \textbf{6.30} & \textbf{59.66} & \textbf{4.93} & \textbf{17.81} \\ 
\bottomrule
\end{tabular}
\label{tab:mob_distill}
\end{table}
These results suggest that ClassroomKD has strong potential to enhance performance on compact datasets, even where traditional methods fall short. By selectively leveraging the most effective mentors, ClassroomKD enables optimal knowledge transfer, making it a promising approach for dataset distillation. Additionally, combining ClassroomKD with dataset distillation can be extended to \textbf{continual learning}, where models from previous tasks act as mentors for new tasks. This approach could improve efficiency and performance in larger-scale tasks and real-world scenarios.

\newpage
\section{Classroom Learning Styles Survey}
\label{sec:survey}

We conducted an online survey about learning styles and academic success in the classroom environment, in which forty (40) respondents participated. Most respondents (92.5\%) were 18-45 years old, with 32.5\% self-identifying as students, 22.5\% as teachers or mentors, and 37.5\% identifying as both. This survey aimed to gather insights into the various methods and strategies students and teachers employ to excel in their academic goals. In this appendix, we provide some statistics from the responses we received. These inspired the ClassroomKD approach introduced in the paper. Participation in the survey was voluntary, and participants could withdraw at any time without penalty.

\textbf{Consent form for the survey}
\begin{figure}[ht]
    \includegraphics[width=\linewidth]{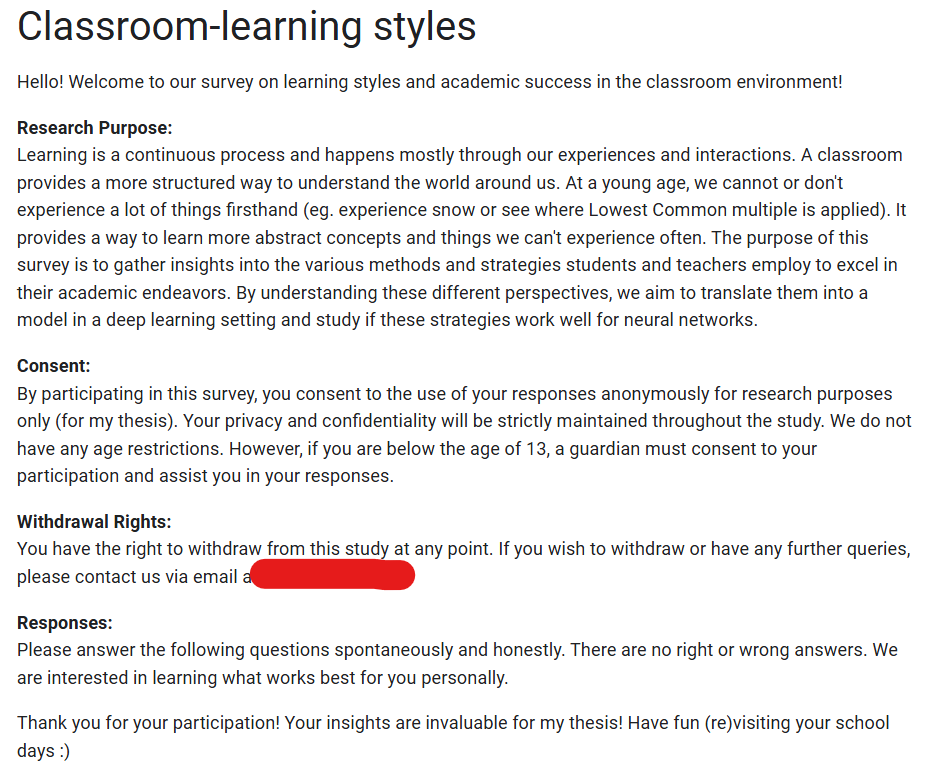}
\end{figure}

\subsection{Role of a Competitive Classroom Environment}

In the first series of questions, we try to find out if students feel like they learn better in collaborative environments, which provide opportunities for healthy competition. The results showed positive response to collaboration among peers along with the teacher. However, competition was mostly detrimental to learning towards the end of the training period (after the completion of coursework and during their exams). 

\textbf{How does competition among peers affect your learning abilities?}
\begin{figure}[ht]
    \includegraphics[width=0.75\linewidth]{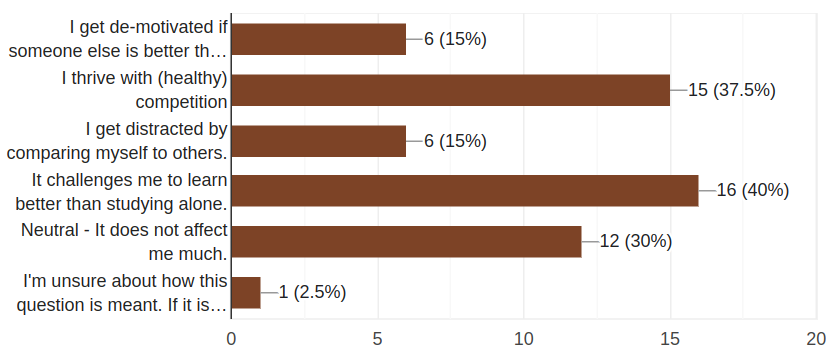}
\end{figure}

The survey further explored specific scenarios where competition was beneficial or detrimental

\textbf{Competition among peers helps me when:}
\begin{figure}[ht]
    \includegraphics[width=0.75\linewidth]{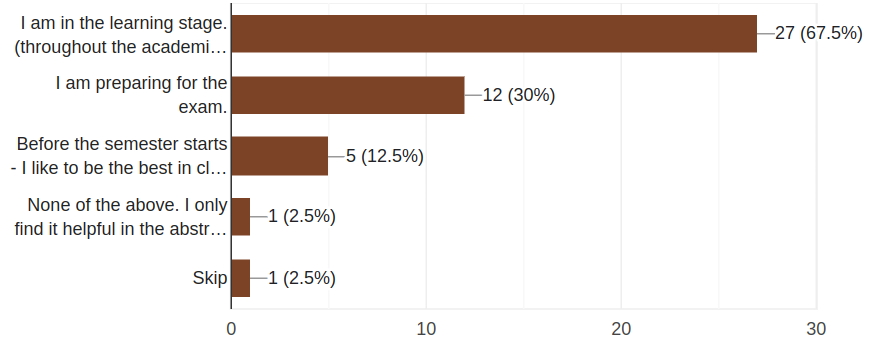}
\end{figure}

Competition was found to be helpful during the learning phase (lecture period) of a semester. This competition can take the form of in-class discussions, group projects, or other collaborative activities. It encouraged active participation and knowledge sharing among students, fostering a collaborative learning atmosphere.

\textbf{Competition among peers is distracting when:}
\begin{figure}[ht]
    \includegraphics[width=0.75\linewidth]{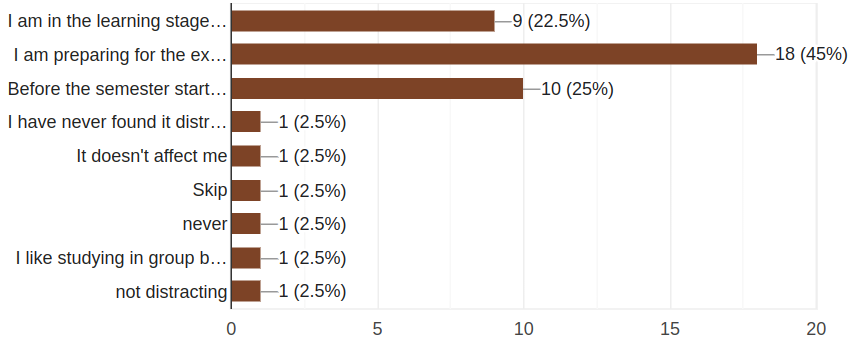}
\end{figure}

On the other hand, competition was often seen as distracting during critical phases like final exams or major project submissions. In these scenarios, the pressure to outperform peers led to decreased focus and increased anxiety, negatively impacting overall performance.

The insights from these responses were instrumental in designing the ClassroomKD framework. Recognizing the dual nature of competition, we incorporated mechanisms to balance collaborative learning with individual performance enhancement:

\begin{itemize}
    \item \textbf{Collaborative Learning Environment}: By integrating multiple peers in the knowledge distillation process, ClassroomKD emulates a collaborative classroom where the student model benefits from diverse feedback. This mirrors the beneficial aspects of peer competition, fostering a supportive learning environment.
    \item \textbf{Performance-Based Filtering}: To mitigate the negative effects of competition, the Knowledge Filtering Module ensures that the student model learns from higher-ranked mentors only. This selective approach reduces the pressure from underperforming models and prevents the error propagation that could arise from unhealthy competition.
\end{itemize}

\subsection{Seeking Guidance}

The second set of questions focused on understanding how students seek guidance when faced with challenges and the effectiveness of the feedback received. In these questions, we attempt to understand what prompts students to seek guidance from their mentors and how they handle it. The goal was to understand the correlation between when or whom students are asking for help and their success in achieving their objectives. 

\textbf{When your confidence drops, whom do you usually ask your doubts?}
\begin{figure}[ht]
    \includegraphics[width=0.7\linewidth]{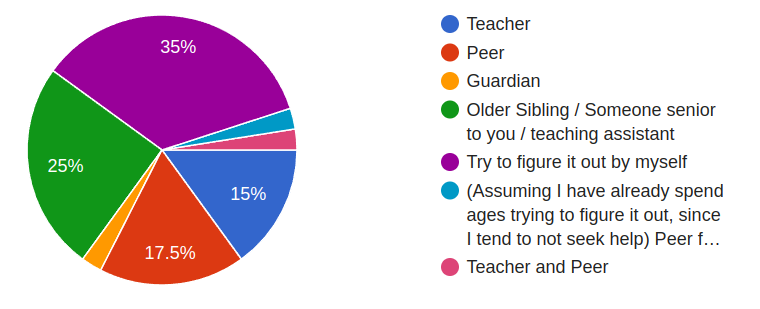}
\end{figure}

The responses indicated a preference for different sources based on the perceived expertise and approachability. Most respondents consulted their peers or older siblings or tried to figure things out themselves. Peers were considered more approachable and could provide relatable explanations.

\textbf{When you asked your questions to your teacher, what was their response?}
\begin{figure}[ht]
    \includegraphics[width=0.7\linewidth]{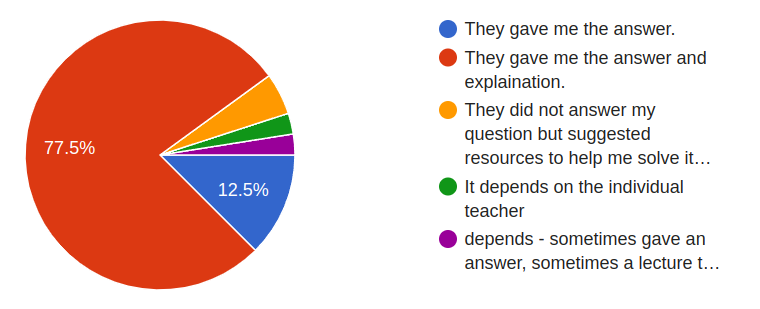}
\end{figure}

When asked about the nature of the teacher's response, many participants noted that teachers often provided detailed explanations and additional resources. This thorough approach helped clarify doubts and improve understanding.

\textbf{Did the teacher's strategy help you gain confidence?}

Many respondents confirmed that their confidence increased after receiving teacher feedback. This highlights the importance of effective mentoring in the learning process.

\begin{figure}[ht]
    \includegraphics[width=0.7\linewidth]{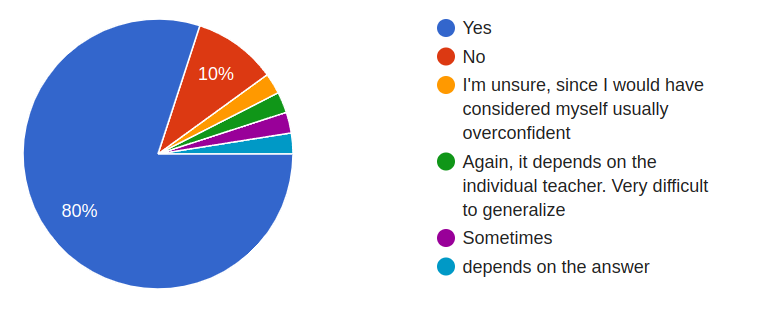}
\end{figure}

These insights were crucial in shaping the Mentoring Module of ClassroomKD:

\begin{itemize}
    \item \textbf{Adaptive Mentoring}: Inspired by the positive impact of teacher feedback, the Mentoring Module dynamically adjusts the teaching strategies based on the student's current performance level. This ensures that the student model receives guidance tailored to its needs, similar to how a teacher would adjust their approach based on a student's understanding.
    \item \textbf{Selective Feedbac}k: To emulate the preference for high-performing peers, the Knowledge Filtering Module ensures that the student model seeks feedback from higher-ranked peers and teachers. This selective process enhances the quality of knowledge transfer and boosts the student model's confidence over time.
\end{itemize}

\subsection{Self-Assessment and Feedback}

The final set of questions aimed to understand how students assess their own performance and the role of feedback in enhancing their learning experience.

\textbf{How do you assess your performance on a test?}
\begin{figure}[ht]
    \includegraphics[width=0.75\linewidth]{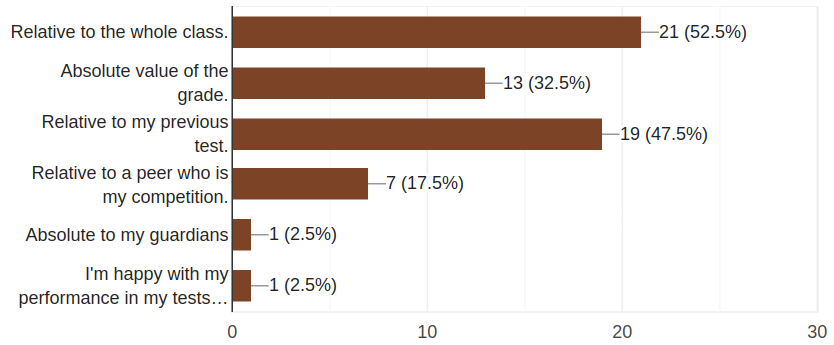}
\end{figure}

Most of the responses suggest that students assess their performance based on peer comparison.

\textbf{My confidence increases when I am appreciated:}
\begin{figure}[ht]
    \includegraphics[width=0.4\linewidth]{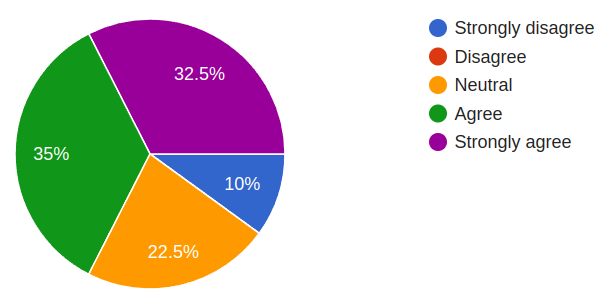}
\end{figure}

Respondents indicated that appreciation from others significantly boosted their confidence. Positive reinforcement motivated them to continue their efforts and strive for better results.

The responses highlighted the importance of self-assessment and constructive feedback, which influenced the design of ClassroomKD:

\begin{itemize}
    \item \textbf{Progressive Confidence Boosting}: Reflecting the impact of appreciation on confidence, ClassroomKD incorporates a Progressive Confidence Boosting strategy. As the student model's performance improves, its self-confidence (represented by the weighting parameter $\alpha$) increases. This dynamic adjustment ensures that the model's learning is reinforced by its achievements, similar to how students gain confidence from positive feedback.
    \item \textbf{Continuous Improvement}: By integrating detailed feedback mechanisms through the Mentoring Module, ClassroomKD ensures that the student model continuously learns from its mistakes. The adaptive teaching strategies help the student model bridge the performance gap with mentors over time, fostering a continuous improvement cycle.
\end{itemize}

The survey responses provided valuable insights into effective learning strategies in a classroom environment. These insights were directly translated into the design and implementation of the ClassroomKD framework, ensuring that our knowledge distillation approach mirrors successful educational practices and optimizes student model performance.

\end{document}